\documentclass[11pt]{article}
\usepackage[preprint]{acl}
\usepackage{times}
\usepackage{latexsym}
\usepackage[T1]{fontenc}
\usepackage[utf8]{inputenc}
\usepackage{microtype}
\usepackage{inconsolata}
\usepackage{graphicx}

\usepackage{multirow}
\usepackage{multicol}
\usepackage{float}
\usepackage{amsmath}
\usepackage{booktabs}
\usepackage{CJKutf8}
\usepackage{dsfont}
\usepackage{xcolor}

\usepackage{xcolor,colortbl}
\usepackage{xcolor}
\newcommand{\atype}[2]{\colorbox{#1!15}{\textsc{#2}}}
\usepackage{anyfontsize}
\usepackage[table]{xcolor}  
\usepackage{colortbl}
\usepackage{arydshln}

\usepackage{amsmath}
\usepackage{amsfonts}
\usepackage{amssymb}

\usepackage{graphicx}
\usepackage{subcaption}

\usepackage{algorithm}     
\usepackage{amsmath}        
\usepackage{algorithm}
\usepackage{algpseudocode}
\usepackage{titletoc}
\usepackage{tcolorbox}   
\usepackage{float}       
\usepackage{verbatim}    
\usepackage{enumitem}
\usepackage{tcolorbox}
\usepackage{pifont}
\usepackage{caption}

\usepackage{fontawesome5} 
\usepackage{hyperref}

\newcommand{\cmark}{\ding{51}} 
\newcommand{\xmark}{\ding{55}} 
\usepackage{enumitem}
\definecolor{amber}{RGB}{255,191,0}
\newcommand{\exper}[1]{\textsc{#1}}
\usepackage{marvosym} % 

\title{Stable-RAG: Mitigating Retrieval-Permutation-Induced \\Hallucinations in Retrieval-Augmented Generation}

\author{
  \textbf{Qianchi Zhang\textsuperscript{1,2}}, 
 \textbf{Hainan Zhang\textsuperscript{1,2}}\textsuperscript{\Letter}\thanks{\textsuperscript{\Letter}Corresponding author.}, 
  \textbf{Liang Pang\textsuperscript{4}}, 
  \textbf{Hongwei Zheng\textsuperscript{3}}, 
  \textbf{Zhiming Zheng\textsuperscript{1,2}}
  \\ 
  \textsuperscript{1}Beijing Advanced Innovation Center for Future Blockchain and Privacy Computing \\
   \textsuperscript{2}School of Artificial Intelligence, Beihang University, China \\
  \textsuperscript{3}Beijing Academy of Blockchain and Edge Computing, China \\
  \textsuperscript{4}Institute of Computing Technology, Chinese Academy of Sciences, Beijing, China \\ 
 \texttt{\{zhangqianchi, zhanghainan\}@buaa.edu.cn}
 }

\makeatletter
\def\thanks#1{\protected@xdef\@thanks{\@thanks
        \protect\footnotetext{#1}}}
\makeatother

\begin{document}
\maketitle
\begin{abstract}

Retrieval-Augmented Generation (RAG) has become a key paradigm for reducing factual hallucinations in Large Language Models (LLMs), yet little is known about how the order of retrieved documents affects model behavior. We empirically show that under a Top-5 retrieval setting with the gold document included, LLM answers vary substantially across permutations of the retrieved set, even when the gold document is fixed in the first position. This reveals a previously underexplored sensitivity to retrieval permutations. Although existing robust RAG methods focus primarily on enhancing LLM robustness to low-quality retrieval and mitigating positional bias to distribute attention fairly over long contexts, neither approach directly addresses permutation sensitivity. In this paper, we propose \textbf{Stable-RAG}, which exploits permutation sensitivity estimation to mitigate permutation-induced hallucinations. Stable-RAG runs the generator under multiple retrieval orders, clusters hidden states, and decodes from a cluster-center representation that captures the dominant reasoning pattern. It then uses these reasoning results to align hallucinated outputs toward the correct answer, encouraging the model to produce consistent and accurate predictions across document permutations. Experiments on three QA datasets show that Stable-RAG improves answer accuracy, reasoning consistency, and generalization across datasets, retrievers, and input lengths compared with strong baselines~\footnote{\textbf{\faGithub~}Our code is available at \url{https://github.com/zqc1023/Stable-RAG}.}.

\end{abstract}

\section{Introduction}

\begin{figure}[!t]
\centering
  \includegraphics[width=1.0\columnwidth]{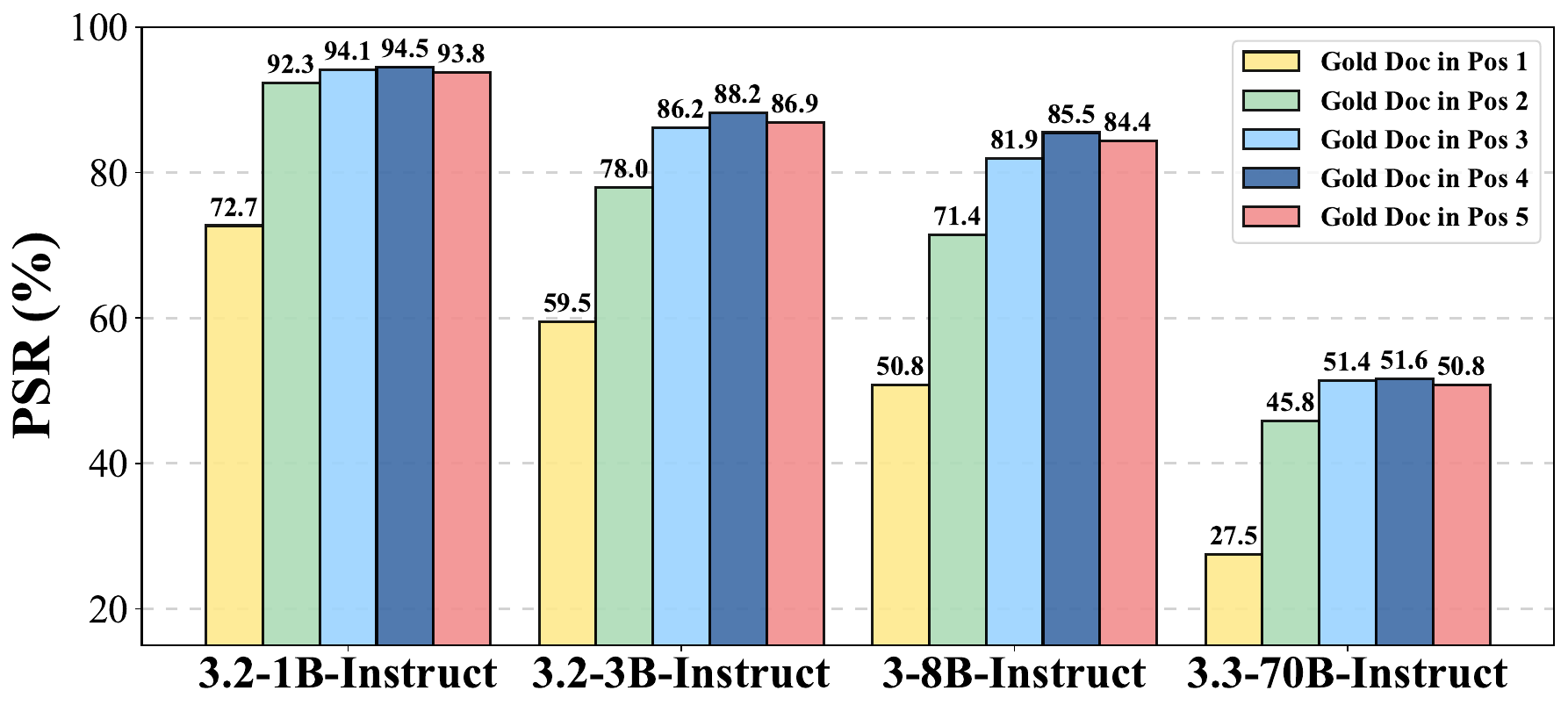}
\caption{\textbf{Perturbation Success Rate (PSR)} on the NQ test set across different LLaMA models. PSR is computed as the proportion of successful document-order perturbations to produce hallucination results among 1,000 randomly sampled instances, with the gold document fixed in different positions. Qwen models' results can be seen in Appendix~\ref{Qwen3}.}
  \label{fig:motivation}
\end{figure}

Large Language Models (LLMs) have achieved remarkable performance on language understanding and generation tasks, but still often generate confident yet incorrect statements, known as hallucinations~\cite{fan2024survey,xu2026selfcorrectingragenhancingfaithfulness}, especially in knowledge-intensive settings~\cite{chen2022gere,huang-etal-2023-learning,zhang2024trendfact,chen2025privacy, luo-etal-2025-dtcrs,li-etal-2025-misleading,yuan2025understanding,ji2026strideedstrategygroundedstepwisereasoning,hu2026contextagent}. Retrieval-Augmented Generation (RAG)~\cite{lewis2020retrieval,gao2023retrieval,li2026modeling} reduces factual hallucinations by grounding model outputs in externally retrieved documents rather than relying only on parametric knowledge, improving factuality, interpretability, and updatability without additional retraining~\cite{zhou2024trustworthiness}.

Despite these benefits, RAG systems are far from hallucination-free~\cite{hamman2025improving}. We identify a critical but overlooked vulnerability in existing RAG systems: a strong sensitivity to the order of retrieved documents.
When the retrieved content remains the same, with the gold document included, merely reordering them can lead the model to follow entirely different reasoning paths and produce inconsistent answers, referred to as \textbf{Permutation-Induced Hallucinations}.
As shown in Figure~\ref{fig:motivation}, we retrieve the Top-5 documents~\cite{xu2024recomp,zhu-etal-2024-information}, place the gold document in different positions, and observe that LLM answers vary substantially across retrieval permutations. Even when the gold document is fixed first, models may ignore it and produce answers that conflict with the evidence.
This reveals a previously underexplored sensitivity to retrieval permutations, even in short contexts under 1,000 tokens.

Existing robust RAG methods mainly focus on retrieval quality and positional bias. The former enhances LLM robustness to low-quality retrieval via uncertainty estimation and adversarial training, such as noise injection~\cite{raat,retrobust} of weakly relevant documents. The latter alleviates attention bias toward specific positions in long contexts, promoting more balanced use of retrieved documents~\cite{mspoe,wang-etal-2025-position}. However, these approaches overlook a critical issue: permutation sensitivity is neither caused by weakly relevant documents, because the input documents are the same, nor confined to long-context reasoning tasks, since only the Top-5 documents fall within one thousand tokens.

\begin{figure}[t!]
    \centering
    \begin{subfigure}{0.49\linewidth}
        \includegraphics[width=\linewidth]{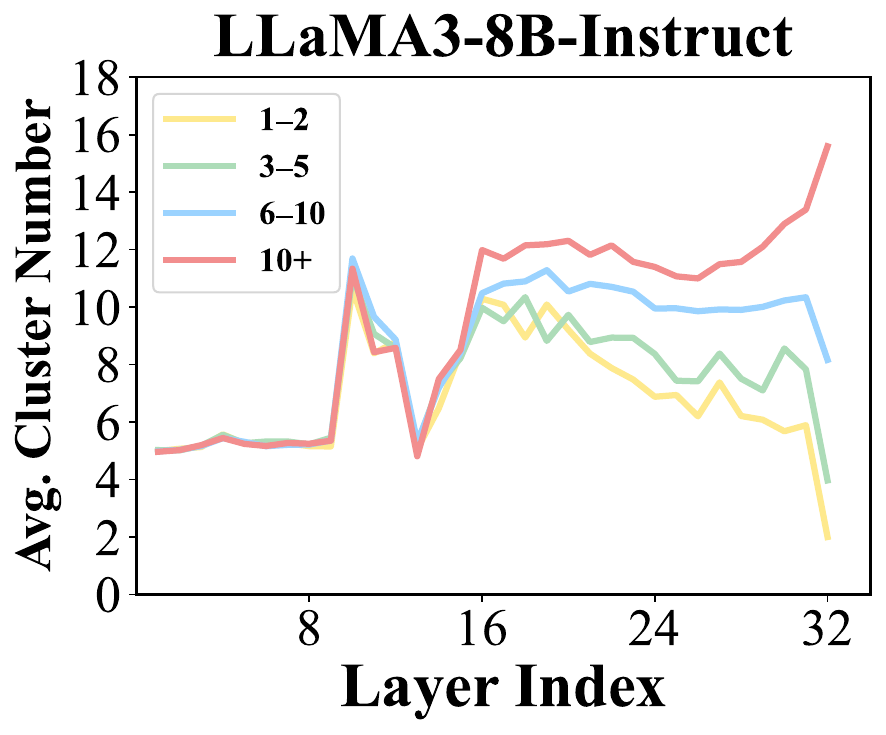}
        % \subcaption{}
        \label{}
    \end{subfigure}
    \hfill
    \begin{subfigure}{0.49\linewidth}
        \includegraphics[width=\linewidth]{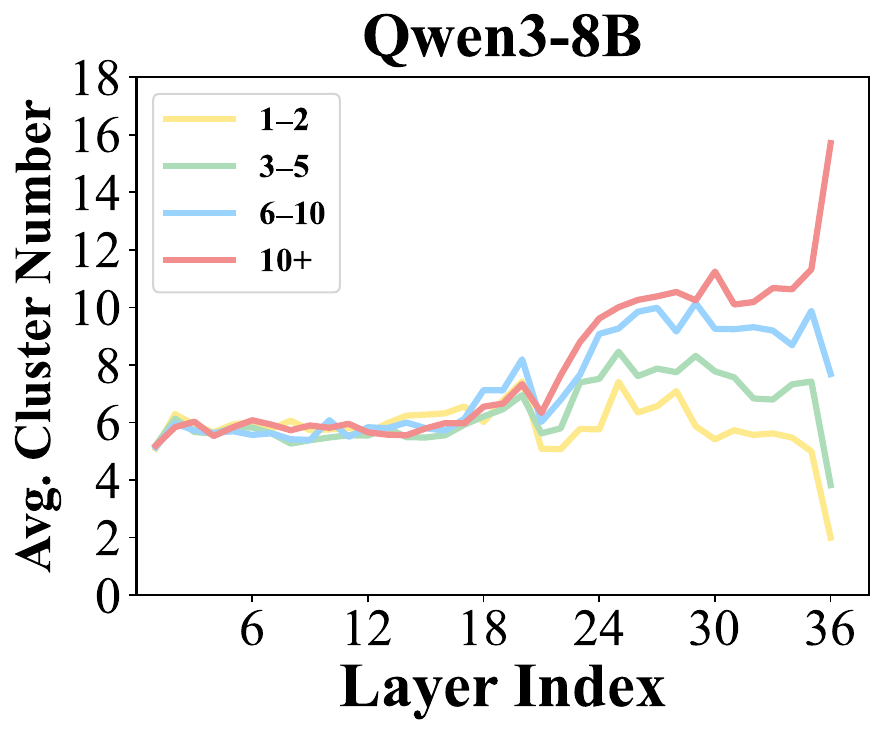}
        % \subcaption{}
        \label{}
    \end{subfigure}
      \caption{Hidden-state clustering behaviors across layers for LLaMA3-8B-Instruct on the NQ train set with DPR retriever and Qwen3-8B on the HotpotQA train set with Contriever retriever, using 1,000 randomly sampled instances. Different colored lines indicate the number of clusters of final reasoning states produced by the LLM under all $5! (=120)$ permutations of the Top-5 retrieved documents (e.g., the green line indicates 3–5 cluster states). Other scales are reported in Appendix~\ref{Instability}.
      }
   \label{fig:hidden_state_compare}
\end{figure}

Instead, permutation sensitivity stems from structural instability in the internal reasoning dynamics of LLMs. As model depth increases, document permutations induce a growing number of distinct reasoning trajectories, leading to more frequent branching and a higher risk of hallucinations or unreliable outputs. As shown in Figure~\ref{fig:hidden_state_compare}, we measure the average number of clusters obtained via spectral clustering over document-permuted representations across different LLM layers on the NQ and HotpotQA datasets. The results indicate that reasoning trajectories in shallow layers are relatively concentrated, while divergence emerges in the middle layers and becomes more pronounced in higher layers. Furthermore, sensitive samples (i.e., 10+) exhibit substantially greater divergence than non-sensitive ones (i.e., 1-2), with this effect primarily localized to the higher layers. These findings highlight the importance of mitigating permutation sensitivity, enabling LLMs to produce stable and accurate outputs regardless of the ordering of retrieved documents, which is critical for improving the robustness of RAG systems.

In this paper, we introduce \textbf{Stable-RAG} that explicitly leverages permutation sensitivity estimation to mitigate the permutation-induced hallucinations.
Specifically, we apply spectral clustering to the last token hidden states of the final layer before response generation, across all document permutations to identify dominant reasoning clusters.
For each cluster, we select a representative hidden state and decode it to obtain candidate answers, thereby capturing the model’s core reasoning modes. Then, we perform cross-cluster consistency alignment over these candidates, encouraging the model to prioritize semantically consistent and factually correct answers across different document orders. This cluster-based alignment substantially reduces the uncertainty induced by order perturbations and fundamentally improves the robustness of RAG.

Experiments on three QA datasets demonstrate that Stable-RAG improves answer accuracy, reasoning consistency, and generalization across datasets, retrievers, and input lengths compared with strong baselines.

Our main contributions are as follows:
\begin{itemize}[leftmargin=*, itemsep=2pt, topsep=2pt]
    \item We find that RAG systems are highly sensitive to document order, leading to inconsistent reasoning. Layer-wise hidden state clustering reveals divergent reasoning trajectories across layers.   
    \item We propose Stable-RAG, which can mitigate permutation-induced hallucinations, achieving model-agnostic stable reasoning.
    \item Across three QA datasets, Stable-RAG outperforms strong baselines in accuracy and reasoning consistency.
\end{itemize}

\section{Related Work}

RAG mitigates factual hallucinations~\cite{yang2026infact,kong2026causalgazeunveilinghallucinationscounterfactual} in LLMs on
knowledge-intensive tasks by providing explicit evidence from external documents~\cite{lewis2020retrieval, fan2024survey,qian2025hyfedrag,yuan2026strucsum,luo2026agscadaptivegranularitysemantic,ge2026expert}.
Prior work on improving the robustness of RAG systems has primarily focused on enhancing retrieval quality~\cite{liu2021distilling,xu2024recomp,ma-etal-2024-context,li2026retrievalgenerationunifiedframework} and reranking performance~\cite{liu-etal-2021-improving-embedding-based,liu2025queries}, or strengthening the generator’s robustness. For instance, AdaComp~\cite{zhang2024adacomp} and CompSelect~\cite{compselect} apply noise filtering to boost generation accuracy. RetRobust~\cite{retrobust} and RAAT~\cite{raat} expose the model to retrieval noise or irrelevant documents during training, enhancing robustness.
However, these methods generally assume a stable document order and do not systematically assess its impact on reasoning. 
Although ATM~\cite{zhu-etal-2024-atm} considers order perturbations, it does not explicitly model reasoning trajectories across permutations and thus cannot ensure consistency.

In addition, another line of research focuses primarily on positional bias in long-context scenarios. Most LLMs use relative positional encodings~\cite{peysakhovich2023attention}, such as RoPE~\cite{su2024roformer} or ALiBi~\cite{press2021train}, which introduce systematic biases: early tokens receive excessive attention due to attention sinks~\cite{xiaoefficient,guattention}, while long-range decay favors recent tokens. Prior work mitigates these issues by modifying positional encodings~\cite{mspoe,chen-etal-2024-fortify,lin2024mixture,egressy2025set}, adjusting causal masks, reweighting attention or hidden states~\cite{hsieh-etal-2024-found}, or using Pos2Distill~\cite{wang-etal-2025-position} to distill knowledge from advantageous to less favorable positions to promote fair attention across tokens. These methods focus on long contexts and do not explicitly address reasoning inconsistency induced by different permutations of the same document set.

\section{Preliminary Study}

\subsection{Problem Formulation}

Given a query $q$ and its retrieved document set $\mathcal{S} = \{d_1, d_2, \dots, d_n\}$, the goal is to ensure that the model $f_\theta$ produces consistent outputs across different document orderings. Let $\mathrm{Perm}(\mathcal{S})$ denote all possible permutations of $\mathcal{S}$. For any two permutations $\pi_1, \pi_2 \in \mathrm{Perm}(\mathcal{S})$, the model’s outputs should be as similar as possible:
\[
f_\theta(q, \pi_1) \approx f_\theta(q, \pi_2).
\]
In this task, the model is expected to produce consistent answers regardless of the document order.

\subsection{Permutation Sensitivity Estimation}

Recent work~\cite{liang2025clue,lee-etal-2025-efficient-latent} exploits hidden states to uncover latent reasoning trajectories, often as indicators of generative uncertainty. Accordingly, we propose to quantify model generation uncertainty via the spectral clustering of hidden states. In this section, we validate the feasibility of the spectral clustering~\cite{ng2001spectral,von2007tutorial} through both layer-wise visualization and quantitative analysis.

\begin{figure}[!t]
\centering
  \includegraphics[width=0.95\columnwidth]{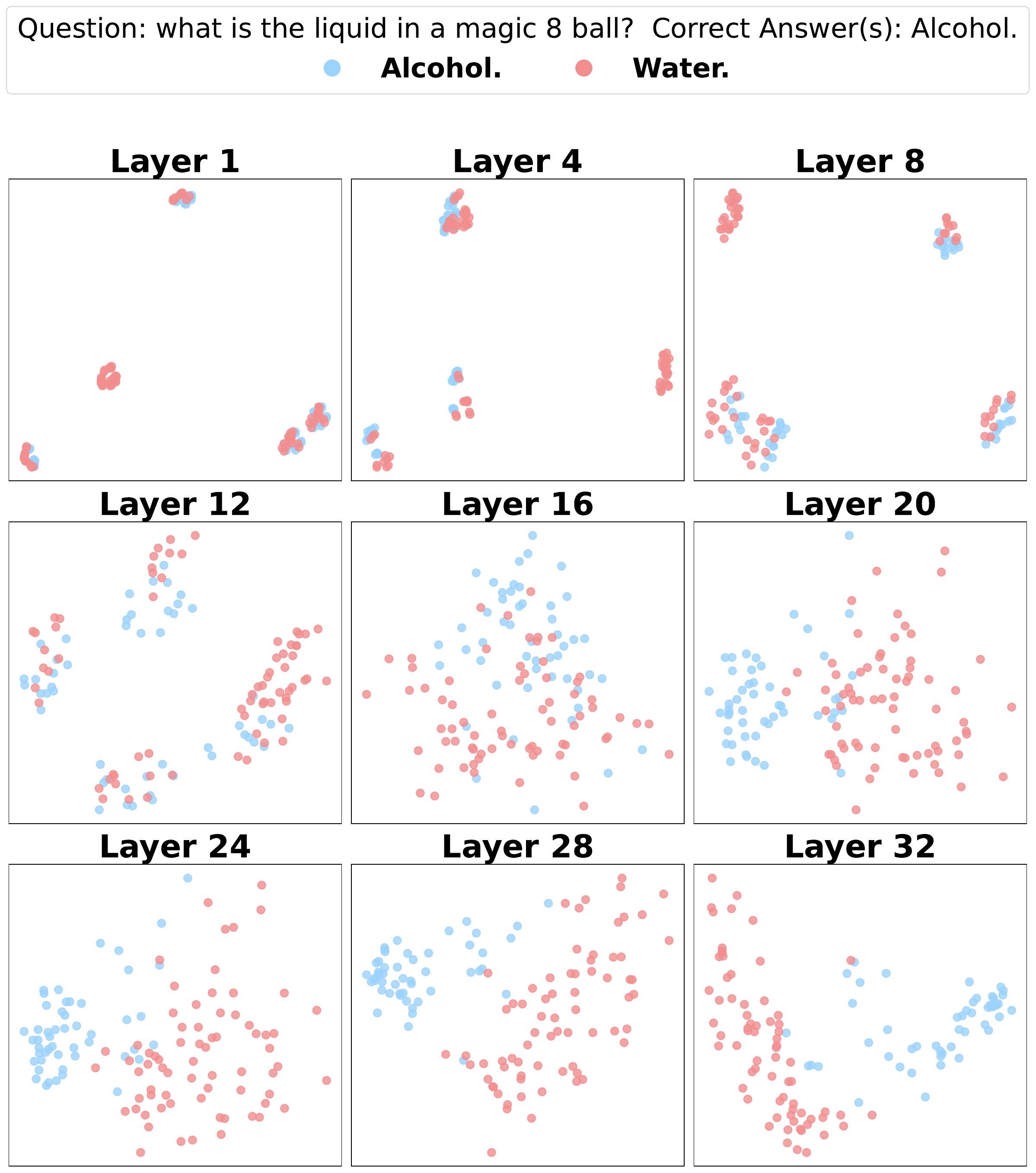}
\caption{Layer-wise visualization of a case study from the NQ train set using LLaMA-3-8B-Instruct. Each point corresponds to a document order, and its color represents the model’s final answer.}
  \label{fig:preliminary}
\end{figure}

\begin{figure*}[!t]
  \centering
  \includegraphics[width=0.96\textwidth]{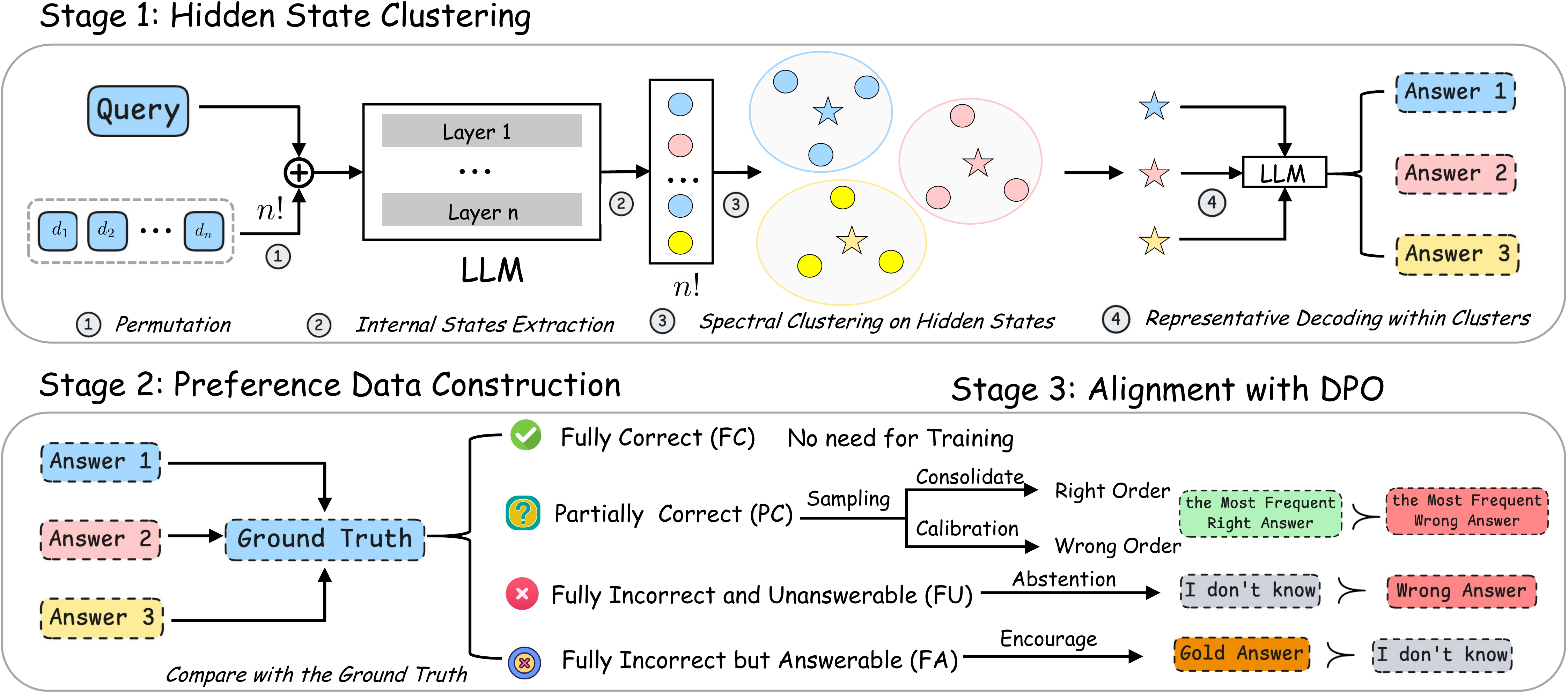}
  \caption{Overall framework of our Stable-RAG.}
  \label{fig:architecture}
\end{figure*}

\paragraph{Layer-wise Visualization.}
For each question, we permute the Top-5 documents to generate $5!=120$ orders and extract the hidden states of the last token from each layer before response generation. Representative layers are then projected to two dimensions via PCA for visualization, as shown in Figure \ref{fig:preliminary}. We observe that hidden states in shallow layers form mixed clusters with points corresponding to different answers interleaved, while in deeper layers the clusters become increasingly well-separated and points with the same answer clearly group together. This indicates that variations in document order induce distinct reasoning trajectories, which manifest as progressively separable clusters in hidden state space, reflecting the model’s internal reasoning patterns. More results are presented in Appendix~\ref{app:all_layer}.

\paragraph{Quantitative Analysis of Clustering.}

\begin{table}[!t]
\centering
\small
\setlength{\tabcolsep}{5pt}
\resizebox{0.95\linewidth}{!}{
\begin{tabular}{l c c c c}
\toprule[1.3pt]
\textbf{Model} & \textbf{Layer} & \textbf{Precision} & \textbf{Recall} & \textbf{F1} \\
\midrule

  \multirow{3}[4]{*}{\textsc{Qwen3-8B}} & 8  & 78.1 & 79.3 & 77.9 \\

& 16 & 79.9 & 81.3 & 79.6 \\
 & 24 & 86.8 & 87.5 & 86.6 \\
 & 36 & 87.8 & 88.4 & 87.6 \\
\midrule
 \multirow{2}[3]{*}{\textsc{LLaMA3-}}& 8  & 69.2 & 71.8 & 69.3 \\
 \multirow{2}[3]{*}{\textsc{8B-Instruct}}& 16 & 81.4 & 82.5 & 81.3 \\
& 24 & 82.3 & 83.7 & 82.2 \\
 & 32 & 84.1 & 85.2 & 83.9 \\
\bottomrule[1.3pt]
\end{tabular}
}
\caption{Clustering performance (\%) of hidden states across different layers for Qwen3-8B and LLaMA3-8B-Instruct on the NQ train set using DPR retriever, averaged over 10,000 randomly sampled instances.}
\label{tab:layer_clustering_results}
\end{table}

To assess each cluster’s reasoning performance, we select the hidden state closest to the cluster center, decode it as a representative answer of the cluster, and match this answer with the real reasoning answers of all hidden states in the same cluster to compute overall Precision, Recall, and F1 scores. As shown in Table~\ref{tab:layer_clustering_results}, clustering metrics improve with network depth, indicating that hidden states for different answers become more separable in deeper layers. 
Notably, the clustering performance is already satisfactory for practical use, with F1 scores of 83.9 using LLaMA3 and 87.6 using Qwen3, respectively. Thus, we use the final layer hidden states for spectral clustering in our method.

\section{Methodology}

\paragraph{Overview.}
Our method comprises three stages: hidden state clustering, preference data construction and alignment with DPO, as shown in Figure~\ref{fig:architecture}. For each permutation, we extract the last token hidden state of the final layer before response generation, capturing the model’s reasoning states. Spectral clustering is then applied to uncover latent reasoning modes, and representative states from each cluster are decoded. By aligning hidden states across permutations, our approach improves generation consistency across different retrieval orders.

\subsection{Hidden State Clustering}

\paragraph{Internal States Extraction.}
For each query $q$ and its retrieved document set $\mathcal{S} = \{d_1, d_2, \dots, d_n\}$, we enumerate all permutations of the documents and run the model for each permutation. Let $i \in \{1, \dots, N\}$ denote the permutation index, where $N = n!$. To reduce computational cost, we extract only the last token hidden state of the final layer before response generation, $h^{(i)} \in \mathbb{R}^d$. Prior work~\cite{azaria-mitchell-2023-internal,ni-etal-2025-towards} has shown that this hidden state sufficiently captures the model’s perception of its knowledge boundaries. We organize all hidden states into a matrix $H$:
\[
H = [h^{(1)}, h^{(2)}, ..., h^{(N)}]^\top \in \mathbb{R}^{N \times d},
\]
which represents the distribution of the model's final reasoning states across document permutations.

\paragraph{Spectral Clustering on Hidden States.}

To determine the number of clusters adaptively and capture the global structure of the hidden state space, we apply spectral clustering~\cite{ng2001spectral} to $H$, where each cluster corresponds to a latent reasoning mode~\cite{lee-etal-2025-efficient-latent}.  
We compute the similarity between each pair of hidden states $h^{(i)}$ and $h^{(j)}$ using the exponential of the cosine distance:
\[
A_{ij} = \exp\Big(-\frac{1 - \frac{h^{(i)} \cdot h^{(j)}}{\|h^{(i)}\| \, \|h^{(j)}\|}}{\sigma}\Big),
\]
where $\sigma$ is a hyperparameter controlling sensitivity. Here, $A \in \mathbb{R}^{N \times N}$ denotes the weighted adjacency matrix of all $N$ hidden states.

The normalized graph Laplacian $L$ is then constructed as:
\[
D = \mathrm{diag}\Big(\sum_{j=1}^N A_{ij}\Big), \quad
L = I - D^{-1/2} A D^{-1/2},
\]
where $D$ is the degree matrix, with each diagonal entry $D_{ii}$ representing the sum of edge weights connected to the $i$-th hidden state (treated as a graph node), and $I$ is the identity matrix.

The number of clusters $K$ is determined adaptively via the eigengap of $L$. Let $\lambda_1 \le \dots \le \lambda_N$ be the eigenvalues of $L$, and define the consecutive gaps $\mathrm{gap}_i = \lambda_{i+1} - \lambda_i$ between each pair of adjacent eigenvalues. The number of clusters is then set as $K = \max\bigl(2, (\displaystyle\arg\max_i \mathrm{gap}_i) + 1\bigr)$ to ensure clear separation between latent reasoning modes.
% following~\citet{von2007tutorial}.
Once $K$ is determined, we obtain normalized spectral embeddings for all hidden states and assign each $h^{(i)}$ to one of the clusters $C_1, C_2, \dots, C_K$. See more details in Appendix~\ref{math}.

\paragraph{Representative Decoding within Clusters.}

Within each cluster $C_k$, we identify a representative hidden state through centroid-based sampling. The cluster centroid is computed as: 
\[
\mu_k = \frac{1}{|C_k|} \sum_{h^{(i)} \in C_k} h^{(i)}.
\]

We select the representative hidden state:
\[
h^{(r_k)} = \arg\min_{h^{(i)} \in C_k} \|h^{(i)} - \mu_k\|_2.
\]

Only the representative hidden states selected within each cluster
$\{h^{(r_1)}, h^{(r_2)}, \ldots, h^{(r_K)}\}$ 
are decoded into textual answers, reducing the number of runs from $N=n!$ to $K$ and substantially lowering computational and annotation overhead.

\paragraph{Exhaustive Full-Permutation Decoding.}
We study an exhaustive permutation decoding setting in which the model is evaluated under all ($N=n!$) permutations of retrieved documents. While this fully characterizes permutation-induced output variability, it is computationally and annotationally prohibitive at scale. We therefore use it only as a reference to assess the efficiency gains of our representative decoding strategy.

\subsection{Preference Data Construction}
\paragraph{Targets.}
% Our goal is to build a robust RAG system. When the model cannot answer, it is encouraged to abstain to suppress hallucinations. When an answer is available, the output should remain consistent regardless of document order, reducing permutation sensitivity.

Our goal is to build a robust RAG system. When the model cannot produce a reliable answer, it is encouraged to abstain to effectively suppress hallucinations and improve system reliability. When an answer is available, the output should remain consistent regardless of document order, thereby reducing permutation sensitivity and further enhancing overall reasoning robustness.

\paragraph{Data Construction Procedure.}
We construct preference data $\mathcal{P}=(x, y_w, y_l)$ for training. For each query $q$ with its retrieved documents set $\mathcal{S}=\{d_1, d_2,\dots,d_n\}$, the input $x$ is formed by concatenating $q$ with a specific document permutation $\pi$. Model outputs are obtained via representative decoding of hidden-state clusters induced by document permutations. Each instance is then compared with the ground truth and categorized into the following four types:
\noindent
\atype{green}{FC} (\textbf{Fully Correct}):
the base model produces correct answers under all document permutations. Such instances are stable and excluded from training.
% , indicating stable behavior for such instances and therefore excluding them from training.
\noindent
\atype{cyan}{PC} (\textbf{Partially Correct}):
the base model produces both correct and incorrect answers across permutations. Two representative outputs are sampled: $y_w$ is the most frequent right answer to consolidate correct predictions, and $y_l$ is the most frequent wrong answer for calibration.
\noindent
\atype{red}{FU} (\textbf{Fully Incorrect and Unanswerable}):
the base model answers incorrectly under all permutations and no gold answers exist in the documents. $y_w$ is set to \emph{``I don’t know''} to encourage abstention, and $y_l$ is the most frequent wrong answer. 
\noindent
\atype{amber}{FA} (\textbf{Fully Incorrect but Answerable}):
the base model answers incorrectly under all permutations but a gold answer exists in the documents. $y_w$ is set to the gold answer to encourage correct prediction, and $y_l$ is \emph{``I don’t know''}.

\begin{table*}[!t]
\centering
\small
\renewcommand\arraystretch{1.1}
\resizebox{\textwidth}{!}{
\begin{tabular}{ l cccc cccc cccc cc}
\toprule[1.3pt]
\multirow{3}[3]{*}{\textbf{Method}} 
& \multicolumn{4}{c}{\textbf{NQ}} 
& \multicolumn{4}{c}{\textbf{TriviaQA}} 
& \multicolumn{4}{c}{\textbf{HotpotQA}}
& \multicolumn{2}{c}{\multirow{2}[2]{*}{\textbf{Average}}}\\

\cmidrule(r){2-5} \cmidrule(r){6-9} \cmidrule(r){10-13} 
& \multicolumn{2}{c}{\textbf{Contriever}} & \multicolumn{2}{c}{\textbf{DPR}}
& \multicolumn{2}{c}{\textbf{Contriever}} & \multicolumn{2}{c}{\textbf{DPR}}
& \multicolumn{2}{c}{\textbf{Contriever}} & \multicolumn{2}{c}{\textbf{DPR}}
&  & \\

\cmidrule(r){2-3} \cmidrule(r){4-5}
\cmidrule(r){6-7} \cmidrule(r){8-9}
\cmidrule(r){10-11} \cmidrule(r){12-13}
\cmidrule(r){14-15}
& SubEM & F1 & SubEM & F1
& SubEM & F1 & SubEM & F1
& SubEM & F1 & SubEM & F1
& SubEM & F1 \\
\hline
 \rowcolor[HTML]{F0F0F0}
   \multicolumn{15}{c}{\textbf{\mbox{\textsc{LLaMA3-8B-Instruct}}}} \\
% \hline
Direct Generation  & 25.18 & 29.11 & 25.18 & 29.11 & 55.92 & 58.95 & 55.92 & 58.95 & 21.39 & 22.87 & 21.39 & 22.87 &34.16&36.98\\

Vanilla RAG  & 40.75 & 42.82 & 45.81 & 47.80  & 63.89 & 65.43& 67.12 & 68.61  & 30.73 & 34.08 & 25.66 & 28.22& 45.66& 47.83\\

Vanilla SFT  & 42.10 & 44.78 & 46.20 & 49.44  & 55.52 & 51.40  & 57.10 & 52.51 & 27.25 & 31.58  & 24.63 & 29.85& 42.13 & 43.26 \\

\cdashline{1-15}
RetRobust  & 41.82 & 44.26 & 48.70 & 49.29  & 64.85 & 66.72 & 68.67 & 70.42  & 31.46 & 35.34 & 26.96 & 30.36 & 47.08 & 49.40 \\
ATM  & 43.75 & 44.88 & 49.78 & 50.19  & 66.37 & 67.12 & 70.12 & 70.35 & 34.36 & 36.97 & 28.55 & 29.31  & 48.82 & 49.80 \\
RAAT   & 42.33 & 43.85 & 49.12 & 49.85  & 65.58 & 66.94 & 68.03 & 69.12  & 33.58 & 36.12 & 26.35 & 28.79 &  47.50 & 49.11 \\
\cdashline{1-15}
Pos2Distill   & 44.58 & 43.12 & 49.25 & 48.37  & 64.13 & 65.78 & 66.57 & 68.12  & 32.73 & 35.79 & 26.45 & 28.91 &47.29&48.35\\
Ms-PoE & 40.32 & 42.49 &45.58 & 47.53   & 64.21 & 66.14 & 66.48 & 67.73  & 30.17 & 33.65 & 26.12 & 28.57& 45.48 & 47.69\\
\cdashline{1-15}
\textbf{Stable-RAG (Ours)}    & \textbf{48.14} & \textbf{45.80} & \textbf{52.02} & \textbf{50.72}  & \textbf{72.05} & \textbf{71.56} & \textbf{73.43} & \textbf{73.76}  & \textbf{38.91} & \textbf{39.87} & \textbf{29.48} & \textbf{31.68} &\textbf{52.34}& \textbf{52.23} \\
\textbf{$\text{Stable-RAG}^{\clubsuit}$ (Ours)}  & \textbf{48.75} & \textbf{46.58} & \textbf{52.88} & \textbf{51.78}  & \textbf{72.13} & \textbf{71.89} & \textbf{74.01} & \textbf{74.12}  & \textbf{39.12} & \textbf{40.16} & \textbf{30.41} & \textbf{32.12} & \textbf{52.88} & \textbf{52.78} \\

\hline
 \rowcolor[HTML]{F0F0F0}
    \multicolumn{15}{c}{\textbf{\mbox{\textsc{Qwen3-8B}}}} \\

Naive Generation & 21.94 & 24.07 & 21.94 & 24.07 & 45.77 & 48.16 & 45.77 & 48.16 & 19.54 & 24.86 & 19.54 & 24.86 & 29.08 & 32.36\\
Vanilla RAG & 44.65 & 45.34 & 50.55  & 50.67  & 64.35 & 66.29  & 69.62 &71.03  & 33.14 & 38.66 & 26.17 & 31.33 & 48.08 & 50.55 \\
Vanilla SFT  & 41.41 & 45.05 & 45.60 & 49.19  & 51.87 & 47.62 & 54.46 & 50.17 & 28.36 & 34.15 & 25.35 & 29.77 &41.18&42.66\\
\cdashline{1-15}
RetRobust  & 43.10 & 44.99 & 49.50 & 50.81  & 63.49 & 65.39 & 69.12 & 70.33 & 32.77 & 39.39& 26.83 & 33.06  &47.47&50.66\\
ATM & 45.47 & 45.86 & 50.94 & 51.03 & 64.78 & 66.57 & 70.06 & 71.67 & 35.12 & 40.69 & 29.07 & 33.43 &49.24&51.54\\
RAAT  & 45.13 & 45.87  & 50.12 & 50.03  & 63.12 & 65.17 & 68.54 & 69.88  & 33.54 & 39.06 & 27.21 & \textbf{33.75}& 47.94 & 50.63\\
\cdashline{1-15}
Pos2Distill  & 44.89 & 45.52 & 50.71 & 50.93 & 64.95 & 66.81 & 69.87 & 71.35 & 33.72 & 39.11 & 26.53 & 31.88 & 48.45 & 50.93 \\
Ms-PoE & 44.39 & 45.12 & 50.04 & 50.08  & 64.88 & 66.72 & 69.03 & 70.84 & 32.98 & 38.21 & 25.93 & 31.02 &47.88&50.33 \\
\cdashline{1-15}

\textbf{Stable-RAG (Ours) }& \textbf{46.12} & \textbf{46.79} & \textbf{51.69} & \textbf{51.78} & \textbf{66.58} & \textbf{68.13 }& \textbf{71.32} & \textbf{72.89} & \textbf{35.73} & \textbf{41.78} & \textbf{30.15} & 33.26 &\textbf{50.27}&\textbf{52.44}\\
\textbf{$\text{Stable-RAG}^{\clubsuit}$ (Ours)} & \textbf{46.94} & \textbf{47.13} & \textbf{52.12} & \textbf{52.38} & \textbf{67.11} & \textbf{68.79}& \textbf{71.74} & \textbf{73.40} & \textbf{36.89} & \textbf{42.94} & \textbf{31.77} & \textbf{35.78} &\textbf{51.10}&\textbf{53.40}\\
\bottomrule[1.3pt]
\end{tabular}
}
\caption{Main results (\%) on three QA benchmarks using two retrievers. $\clubsuit$ denotes our method trained on exhaustive full-permutation decoding.}
\label{tab:main_results}
\end{table*}

\subsection{Alignment with DPO}
We employ Direct Preference Optimization (DPO)~\cite{rafailov2023direct} to train the base model on the constructed preference tuples.  
For each tuple $(x, y_w, y_l)$, DPO maximizes the likelihood of the preferred answer $y_w$ over the less preferred $y_l$:
\[
\begin{aligned}
\mathcal{L}_{\exper{Dpo}}=&-{\mathbb{E}}_{(x, y_w, y_l) \sim \mathcal{D}} \Big[\log \sigma\Big(\beta \log \frac{\pi_\theta(y_w \mid x)}{\pi_\mathrm{ref}(y_w \mid x)} \\
& - \beta \log \frac{\pi_\theta(y_l \mid x)}{\pi_\mathrm{ref}(y_l \mid x)} \Big) \Big],
\end{aligned}
\]
where $\theta$ denotes the model parameters, $\sigma$ is the sigmoid function, and $\beta$ is a scaling hyperparameter controlling the sharpness of preference. The model policy $\pi_\theta$ is initialized using the base reference policy $\pi_\mathrm{ref}$.

\section{Experiments}

\subsection{Experimental Setup}
\paragraph{Datasets.}
We evaluate our method on three QA benchmark datasets, including (1) Open-Domain QA, represented by NaturalQuestions (NQ)~\cite{nq} and TriviaQA~\cite{triviaqa}; (2) Multi-Hop QA, represented by HotpotQA~\cite{hotpotqa}. 
Dataset statistics are provided in Appendix~\ref{appendix:datasets}.

\paragraph{Evaluation Metrics.}
Since answer style mismatch may cause additional variance, we follow prior and concurrent work~\cite{zhu-etal-2024-atm,wang2025maferw,compselect,limol,luo2026gcotdecodingunlockingdeepreasoning} and adopt Substring Exact Match (\textbf{SubEM}) and \textbf{F1} for evaluation. SubEM checks whether the gold answer appears as a substring in the prediction, while F1 measures token-level overlap with the reference.

\paragraph{Baselines.}

We compare our method with the following baseline strategies on the same test set. Vanilla methods include \textit{Direct Generation}, \textit{Vanilla RAG}~\cite{lewis2020retrieval}, and \textit{Vanilla SFT}~\cite{zhang-etal-2024-r}. Robust RAG methods include \textit{RetRobust}~\cite{retrobust}, \textit{ATM}~\cite{zhu-etal-2024-atm}, and \textit{RAAT}~\cite{raat}. Positional Bias methods include \textit{Pos2Distill}~\cite{wang-etal-2025-position} and \textit{Ms-PoE}~\cite{mspoe}. The details of these baselines are presented in Appendix~\ref{appendix:baseline}.

\paragraph{Implementation Details.}
We use LLaMA3-8B-Instruct~\cite{llama3} and Qwen3-8B~\cite{qwen3} as backbone models for experiments.
To ensure high and consistent evaluation quality~\cite{cuconasu2024power} and further assess the stability of our method under different retrieval settings, we follow prior work~\cite{xu2024recomp,zhu-etal-2024-information,compselect} and use the same Top-5 Wikipedia passages retrieved by DPR~\cite{dpr} and Contriever-MS MARCO~\cite{izacard2021unsupervised} for all baselines and our method.
Additional implementation details are provided in Appendix~\ref{appendix:training}.

\begin{figure*}[t!]
    \centering
    \begin{subfigure}{0.325\linewidth}
        \includegraphics[width=\linewidth]{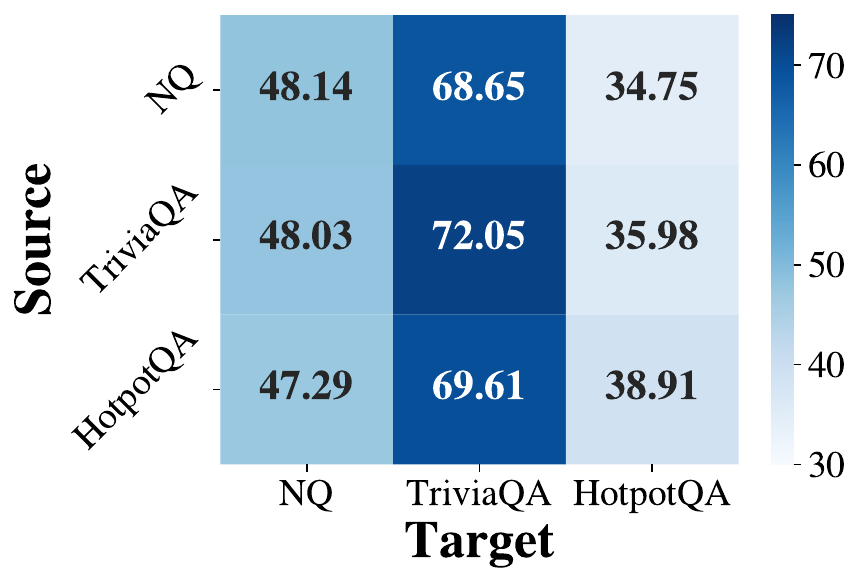}
        \label{fig:ood}
    \end{subfigure}
    % \hfill
    \begin{subfigure}{0.325\linewidth}
        \includegraphics[width=\linewidth]{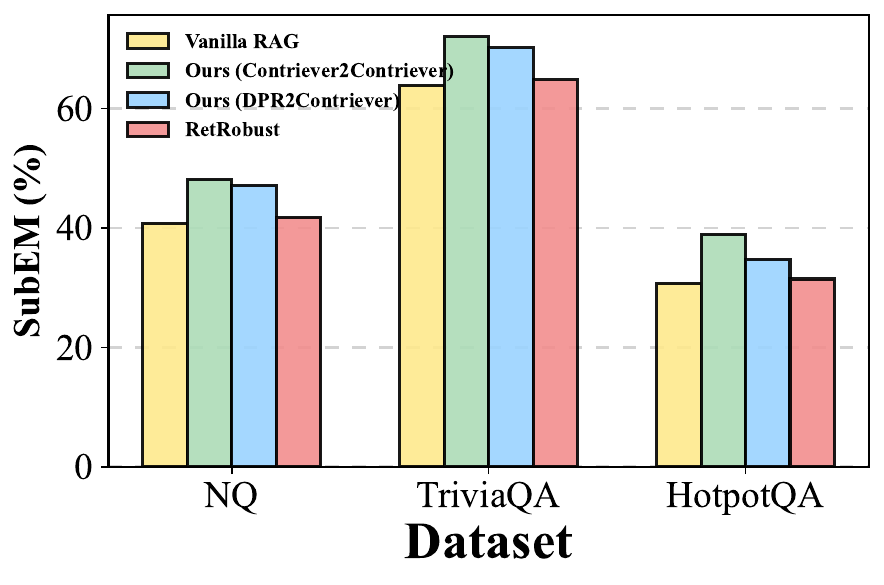}
        \label{fig:retriever}
    \end{subfigure}
    % \hfill
 \begin{subfigure}{0.325\linewidth}
        \includegraphics[width=\linewidth]{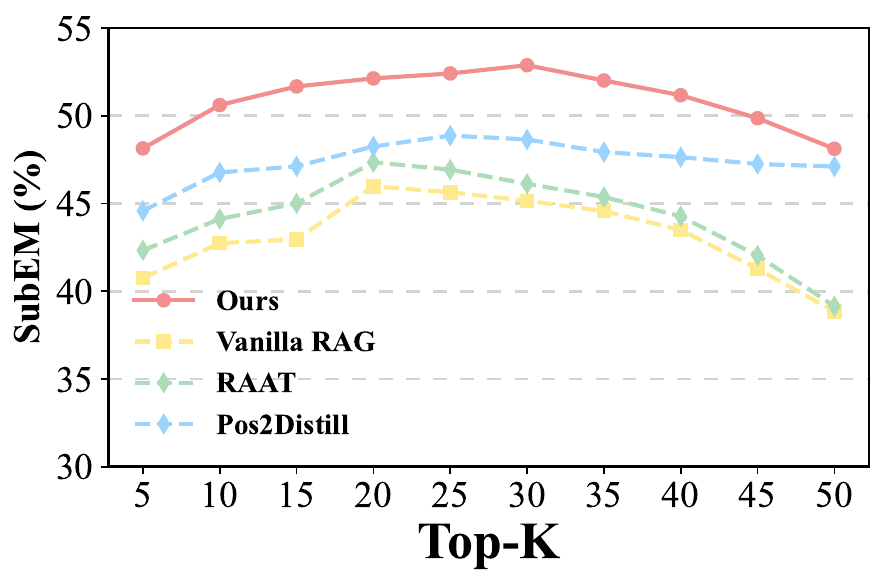}
        \label{fig:topk}
    \end{subfigure}
    
\caption{
\textbf{(Left)} \textit{Cross-Dataset Generalization}. We evaluate on three test sets with the Contriever retriever using SubEM.  
\textbf{(Middle)} \textit{Cross-Retriever Transferability}.
\textbf{(Right)} \textit{Cross-Top-K Robustness}. We evaluate on the NQ test set with the Contriever retriever. All the three experiments are conducted on LLaMA3-8B-Instruct.
}
    \label{fig:generalization}
    \vspace{-10pt}
\end{figure*}

\subsection{Main Results}

We conduct a comprehensive comparison of Stable-RAG against all the baseline methods, as shown in Table~\ref{tab:main_results}. The results indicate the following:
(i) \textbf{Overall performance.} \quad Stable-RAG consistently achieves the best overall performance across all the datasets with both Contriever and DPR retrievers, outperforming all strong baselines;
(ii) \textbf{Effectiveness on complex reasoning.} \quad Stable-RAG consistently improves performance on both single-hop and multi-hop QA tasks, demonstrating its ability to stabilize intermediate reasoning for complex questions;
(iii) \textbf{Model generalization.} \quad Stable-RAG performs robustly across backbone models, indicating model-agnostic generalization.

\subsection{Further Analysis}
\label{analysis}
\paragraph{Ablation Study.}
We conduct an ablation study to assess the contribution of each component in Stable-RAG, as shown in Table~\ref{tab:ablation}. Removing any component consistently degrades performance, demonstrating that all components are essential. In particular, excluding the \textit{PC} component (Index a) causes significant drops across datasets, indicating the importance of partially correct signals for stabilizing reasoning. Removing \textit{FA} (Index c) mainly impacts overall performance, while removing \textit{FU} (Index b,d) sharply reduces the abstention rate~\cite{wen-etal-2025-know, sun-etal-2025-causalabstain}, underscoring its role in handling unanswerable or hallucinated cases. Overall, Stable-RAG achieves the best trade-off between performance and abstention.

\begin{table}[t!]
    \centering
    \renewcommand{\arraystretch}{1.1}
     \resizebox{ \columnwidth}{!}{
    \begin{tabular}{c ccc ccc c c}
        \toprule[1.3pt]
         \multirow{2}{*}{\textbf{Index}}& \multicolumn{3}{c}{\textbf{component}} & \multicolumn{3}{c}{\textbf{Dataset}} & \multirow{2}[3]{*}{\textbf{Average}}& \multirow{2}[3]{*}{\textbf{AR}} \\
         \cmidrule(lr){2-4} \cmidrule(lr){5-7}
         &\textit{PC}   & \textit{FA} &\textit{FU} & NQ & TriviaQA & HotpotQA  & &\\
         \midrule
        (a) & \xmark  & \cmark & \cmark  & 37.62 & 61.37 & 28.54 & 42.51 &\textbf{35.1}\\
        (b) & \cmark  & \xmark & \xmark  & \underline{47.17} & \underline{71.28}& 37.44 & 51.96 &0.0\\
        (c) & \cmark & \xmark & \cmark&46.73 & 70.14 & 35.75 & 50.87 &17.3\\
        (d) & \cmark & \cmark & \xmark& 46.70 & 70.69 & \textbf{38.93} & \underline{52.11} &0.5\\
         % \hline
         \rowcolor[HTML]{F0F0F0}
      \textbf{Ours} & \cmark  & \cmark &\cmark  & \textbf{48.14}& \textbf{72.05}& \underline{38.91}& \textbf{53.03} & \underline{21.8}\\
         \bottomrule[1.3pt]
    \end{tabular}}
   \caption{Ablation results (\%) on LLaMA3-8B-Instruct with the Contriever retriever measured by SubEM. \textbf{AR}(Abstention Rate) denotes the proportion of abstentions on 1,000 randomly sampled questions from three datasets when no retrieval evidence is available and the base model cannot answer. Higher AR indicates better awareness of model limitations and evidence availability. Best and second-best results are bolded and \underline{underlined}, respectively.}
     \label{tab:ablation}
\end{table}

\paragraph{Comparison with Standard DPO.}
To isolate the effect of the order-stability mechanism, we compare Stable-RAG with standard DPO using the same base model and optimization strategy, differing only in whether reasoning consistency across document orders is enforced. In standard DPO, the model is trained to prefer the gold answer when evidence is available over other wrong answers obtained via sampling, or \emph{``I don’t know''} when the query is unanswerable. Results in Table~\ref{tab:dpo_comparison} demonstrate that adding the order-stability constraint consistently improves RAG performance across datasets and retrievers without modifying the preference optimization framework.

\begin{table}[!t]
\centering
\renewcommand{\arraystretch}{1.1}
\resizebox{\columnwidth}{!}{
\begin{tabular}{l cc cc cc}
\toprule[1.3pt]
\multirow{2}[2]{*}{\textbf{Method}} 
& \multicolumn{2}{c}{\textbf{NQ}} 
& \multicolumn{2}{c}{\textbf{TriviaQA}} 
& \multicolumn{2}{c}{\textbf{HotpotQA}} \\
\cmidrule(lr){2-3} \cmidrule(lr){4-5} \cmidrule(lr){6-7}
& Contriever & DPR &  Contriever & DPR &  Contriever & DPR \\
\midrule
Standard DPO 
& 44.76 & 50.88
& 68.03& 71.67 
& 35.96 & \textbf{30.43 }\\

\textbf{Ours}
& \textbf{48.14} & \textbf{52.02} 
& \textbf{72.04} & \textbf{73.43} 
& \textbf{38.91} & 29.48 \\

\bottomrule[1.3pt]
\end{tabular}
}
\caption{SubEM results (\%) between our method and Standard DPO using LLaMA3-8B-Instruct.}
\label{tab:dpo_comparison}
\end{table}

\paragraph{Cross-Dataset Generalization.}  
We further evaluate the transferability of Stable-RAG across different data distributions. As shown in Figure~\ref{fig:generalization} (Left), permutation-sensitivity patterns are learned on an in-domain dataset and directly applied to multiple out-of-distribution datasets to assess cross-dataset generalization. Experimental results demonstrate that Stable-RAG exhibits robust transfer across tasks and knowledge domains, consistently outperforming the best baseline regardless of the source–target dataset combination, and achieving stable improvements in answer consistency.

\paragraph{Cross-Retriever Transferability.}
We further evaluate the model’s transferability by training on the DPR retriever and evaluating on the Contriever retriever. Figure~\ref{fig:generalization} (Middle) shows that the model maintains stable performance under cross-retriever settings, demonstrating strong transferability to different retrieval methods. Additionally, the results of training on the Contriever retriever and evaluating on the DPR retriever are shown in Appendix~\ref{app:across-retriever}.

\begin{figure}[t!]
    \centering
    \begin{subfigure}{0.48\linewidth}  %
        \includegraphics[width=\linewidth]{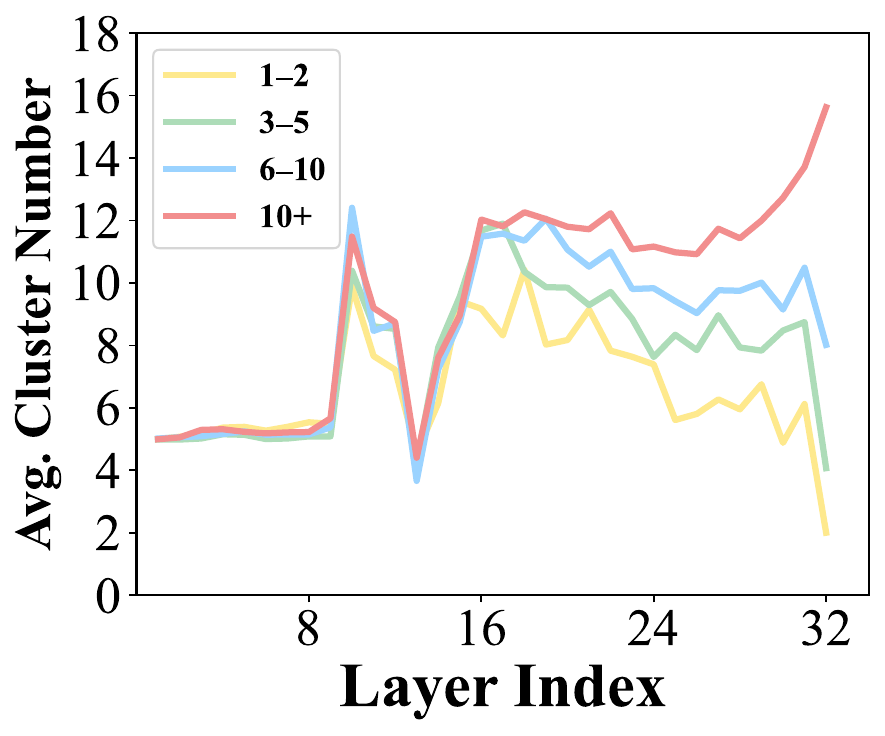}
        \subcaption{LLaMA3-8B-Instruct.}
        \label{base_model}
    \end{subfigure}
    \hfill
    \begin{subfigure}{0.48\linewidth}  
        \includegraphics[width=\linewidth]{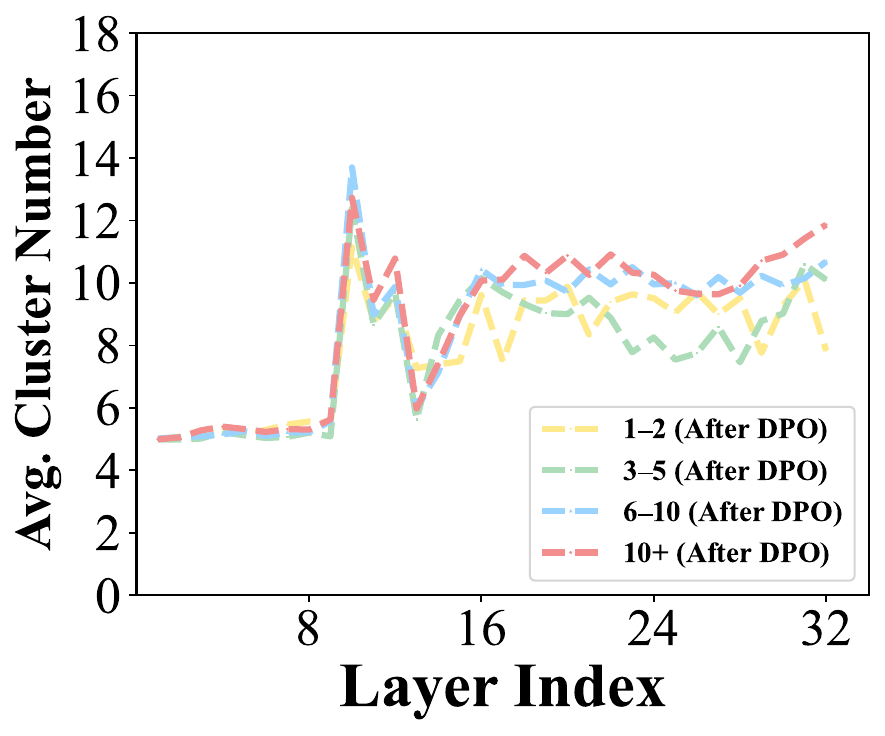}
        \subcaption{Ours.}
        \label{ours}
    \end{subfigure}

    \vspace{0.5em}  

    \begin{subfigure}{0.48\linewidth} 
        \includegraphics[width=\linewidth]{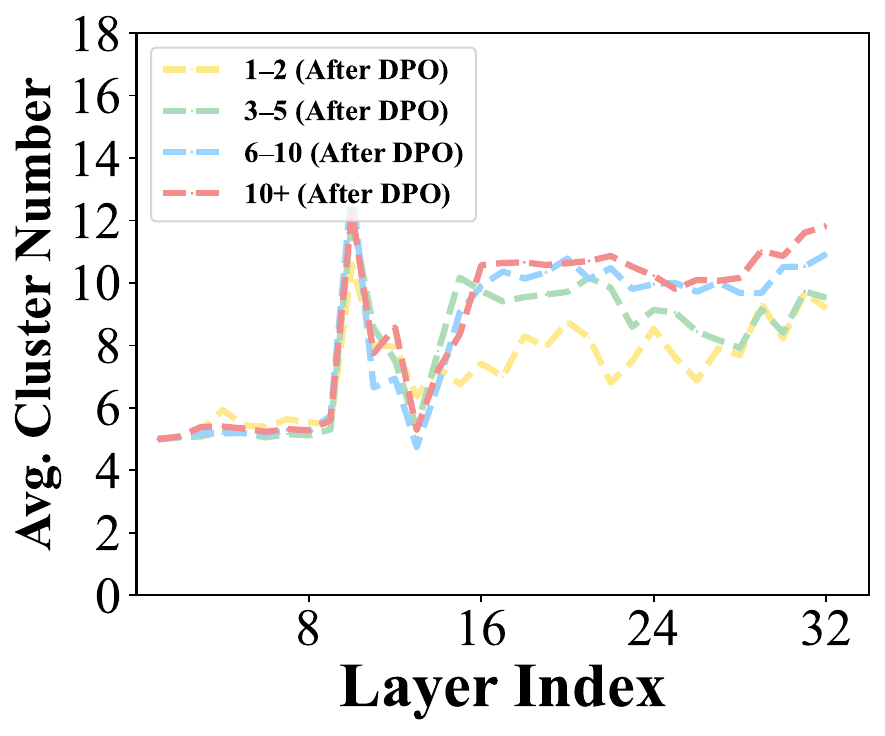}
        \subcaption{Ours (\textit{w/o} \textbf{FU}, \textit{w/o} \textbf{FC}).}
        \label{without_two}
    \end{subfigure}
    \hfill
    \begin{subfigure}{0.48\linewidth}  
        \includegraphics[width=\linewidth]{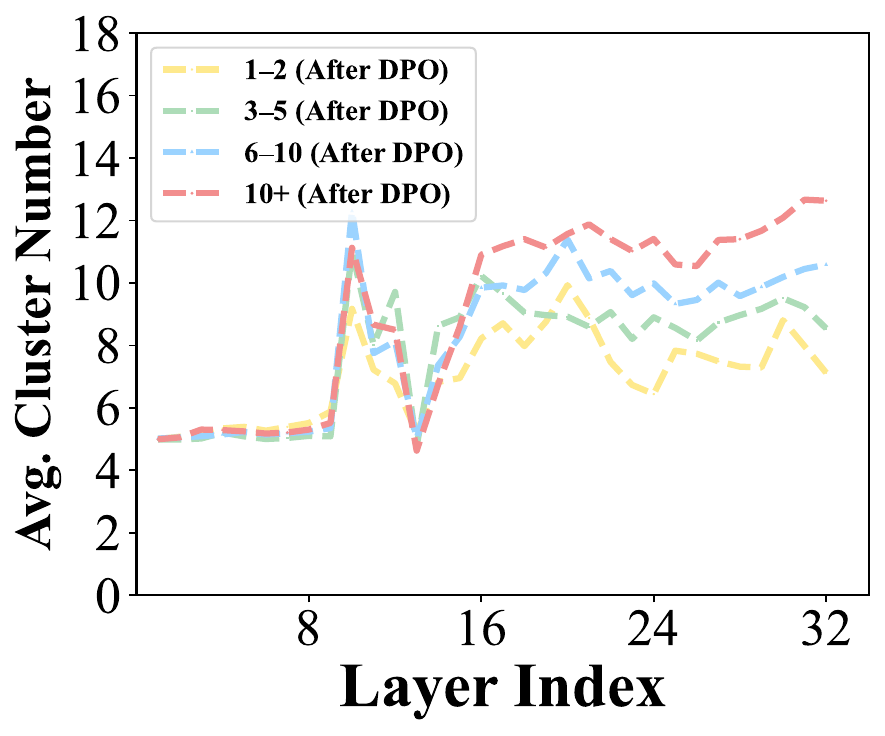}
        \subcaption{Standard DPO.}
        \label{without_pc}
    \end{subfigure}

    \caption{Comparison of internal model behaviors across Base Model (\textbf{a}), Ours (\textbf{b}), one variant of Ours (\textbf{c}), and Standard DPO (\textbf{d}) on a random subset of 500 samples from the NQ test set with Contriever retriever.}
    \label{fig:qwen3_cluster}
\end{figure}
\paragraph{Cross-Top-K Robustness.}

We train the model under a Top-5 setting and evaluate its performance on contexts retrieved with different Top-K values. Experimental results in Figure~\ref{fig:generalization} (Right) show that the model maintains stable performance across various Top-K configurations and achieves significant improvements over corresponding baselines, demonstrating strong generalization when handling different numbers of candidate documents.

\begin{figure}[!t]
\centering
  \includegraphics[width=0.85\columnwidth]{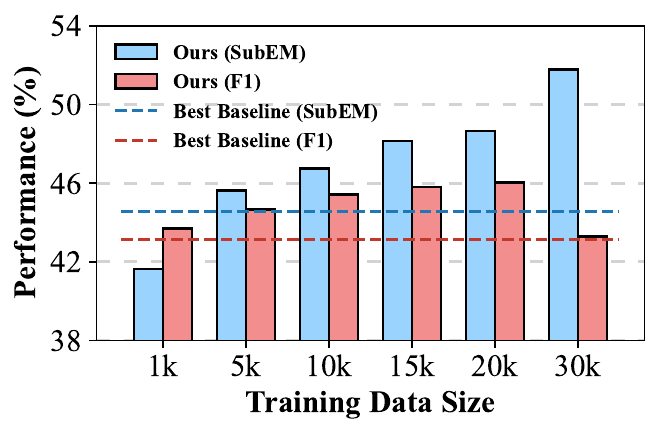}
\caption{Effect of training data size on LLaMA3-8B-Instruct with Contriever retriever on the NQ dataset.}
  \label{fig:training_sample}
\end{figure}

\paragraph{Effect of Training Data Size.}
As shown in Figure~\ref{fig:training_sample}, we analyze the effect of training sample size on learning permutation sensitivity. Performance improves steadily with more data and saturates beyond 15k samples, indicating relatively small datasets suffice to capture core permutation-sensitivity patterns. However, with very limited data (e.g., 1k), performance drops markedly, reflecting difficulty in modeling fine-grained order differences. Given this trade-off, we adopt 15k samples as default, since gains over 20k do not justify the added computational cost.

\paragraph{Internal Model Behaviors after DPO.}  
We label samples by their sensitivity according to the Base Model and exam hidden-state clustering after training. Figure~\ref{ours} shows our method reduces clusters for high-sensitivity samples, keeps medium-sensitivity samples stable, and slightly increases low-sensitivity clusters. Figure~\ref{without_two} shows training on sensitive samples only, and Figure~\ref{without_pc} shows standard DPO results. We can see that the increased clusters mainly stems from DPO-induced answer diversity rather than direct training on sensitive samples. For instance, for the same query \emph{"when was the cat and mouse act introduced?"} and order, the response changes from \emph{"1913."} to “\emph{"introduced in April 1913."} after DPO. Overall, our method stabilizes high-sensitivity representations while preserving diversity for less sensitive samples.

\begin{table}[!t]
\centering
\small
\renewcommand{\arraystretch}{1.1}
\resizebox{1.0\linewidth}{!}{
\begin{tabular}{l c c c c c }
\toprule[1.3pt]
\multirow{2}[2]{*}{\textbf{Method}} & \multicolumn{5}{c}{\textbf{Position of Gold Document}}  \\
\cmidrule(lr){2-6}
 & \textbf{in Pos 1} & \textbf{in Pos 2} & \textbf{in Pos 3} & \textbf{in Pos 4} & \textbf{in Pos 5} \\
\midrule
Vanilla RAG   & 50.8 & 71.4 & 81.9 & 85.5 & 84.4 \\
Vanilla SFT   & 47.2 & 66.2& 74.8 & 80.0 & 82.6 \\
RetRobust     & 35.5 & 75.5& 85.3 & 88.6 & 88.9 \\
ATM           & 33.7 & 64.2& 71.8 & 77.4 & 77.8 \\
Pos2Distill   & 29.5 & 55.8& 69.4 & 72.8 & 73.2 \\
Ms-PoE        & 31.4 & 63.8&  72.1 & 73.9 & 74.3 \\
 \rowcolor[HTML]{F0F0F0}
\textbf{Ours}      & \textbf{28.3} &\textbf{ 54.7} & \textbf{67.3} & \textbf{72.6} & \textbf{73.0} \\
\bottomrule[1.3pt]
\end{tabular}
}
\caption{\textbf{PSR} (\%) on the NQ test set with DPR retriever across different document positions, same as Figure~\ref{fig:motivation}.}
\label{tab:positional_robustness}
\end{table}

\begin{table*}[!t]
\centering
\small
\renewcommand{\arraystretch}{1.05}
\begin{tabular}{l ccc ccc ccc}
\toprule[1.3pt]
\multirow{2}[3]{*}{\textbf{Method}} & \multicolumn{3}{c}{\textbf{NQ}} & \multicolumn{3}{c}{\textbf{TriviaQA}} & \multicolumn{3}{c}{\textbf{HotpotQA}} \\
\cmidrule(lr){2-4} \cmidrule(lr){5-7} \cmidrule(lr){8-10}
 & Original & Shuffled & Drop & Original & Shuffled & Drop & Original & Shuffled & Drop \\
\midrule
Vanilla SFT & 42.10 & 36.43 & 5.67 & 55.52 & 53.19 & 2.33& 27.25 & 22.48  & 4.77\\
RetRobust & 41.82 & 38.06& 3.76& 64.85 & 62.86& 1.99 & 31.46 & 29.18& 2.28\\
ATM & 43.75 & 42.47& 1.28& 66.37 & 63.60& 2.77 & 34.36 &  32.46& 1.90\\
RAAT & 42.33 & 40.54& 1.79& 65.58 & 62.19& 3.39& 33.58 & 29.75& 3.83 \\
Pos2Distill & 44.58 & 43.63& 0.95 & 64.13 & 63.57& 0.56& 32.73 & 32.09& \textbf{0.64}\\
Ms-PoE & 40.32 & 39.17& 1.15 & 64.21 & 62.96& 1.25& 30.17 &  29.14& 1.03\\
\rowcolor[HTML]{F0F0F0}
\textbf{Ours} & \textbf{48.14} & \textbf{47.23} & \textbf{0.91} &\textbf{72.05} & \textbf{71.76} & \textbf{0.29} & \textbf{38.91} & \textbf{37.50} & 1.41\\
\bottomrule[1.3pt]
\end{tabular}
\caption{Performance comparison of LLaMA3-8B-Instruct with Contriever retriever under original and shuffled document order across three QA datasets. We report SubEM for evaluation.}
\label{tab:order_comparison}
\end{table*}

\paragraph{External Positional Robustness after DPO.}
Following prior settings in Figure~\ref{fig:motivation}, we evaluate PSR on 1,000 randomly sampled instances by inserting the gold document in varying positions in the retrieved context to assess external positional robustness using LLaMA3-8B-Instruct. As shown in Table~\ref{tab:positional_robustness}, our method consistently achieves lower PSR across all positions than the baselines, indicating reduced sensitivity to document ordering and improved external robustness under positional perturbations, even when the gold document appears in less favorable or later positions within the context.

\paragraph{Original vs. Shuffled Order}
As shown in Table~\ref{tab:order_comparison}, we present a comparison of answer performance under the original document order and a randomly shuffled order across three QA datasets. Our method achieves the highest SubEM scores in both original and shuffled conditions across all datasets, demonstrating its robustness to retrieval order permutations and its ability to maintain stable answer consistency.

\section{Conclusion}

We identify an underexplored vulnerability in RAG: LLMs are highly sensitive to document order, producing divergent reasoning and inconsistent or hallucinatory outputs from identical evidence. Layer-wise analysis traces this instability to the model’s middle and higher layers. We propose Stable-RAG, which reduces permutation-induced uncertainty by clustering permuted hidden states and aligning reasoning modes via DPO optimization. Experiments across multiple QA benchmarks show consistent gains in accuracy, reasoning stability, and strong transferability. Enforcing layer-wise reasoning constraints while reducing training costs offers a promising approach to mitigate permutation-induced hallucinations.

\section*{Limitations}
While this work demonstrates the effectiveness of Stable-RAG in mitigating permutation-induced hallucinations, it has several limitations that warrant further investigation.

First, our approach focuses on stabilizing reasoning at the final-layer representation level, without explicitly enforcing layer-wise reasoning path constraints throughout the model. Although our analysis reveals that permutation-induced divergence primarily emerges in the middle and higher layers, Stable-RAG does not directly regularize intermediate-layer reasoning trajectories. Incorporating explicit layer-wise constraints or trajectory-level alignment may further improve reasoning stability, but would require more fine-grained supervision or architectural modifications, which we leave for future work.

Second, Stable-RAG relies on spectral clustering over document-permuted hidden representations to estimate dominant reasoning modes and construct preference signals for DPO alignment. While this strategy reduces annotation cost by approximately threefold compared to exhaustive full-permutation decoding, it still incurs non-trivial computational and labeling overhead. More efficient clustering strategies, weak supervision signals, or fully unsupervised alignment objectives could further reduce annotation requirements and improve scalability. 
Exploring such cost-effective supervision mechanisms is important for building more robust and practical RAG systems.

\section*{Ethical Considerations}

While Stable-RAG aims to enhance RAG robustness and accuracy, ethical considerations remain. First, although our method reduces hallucinations from document order, it cannot guarantee fully correct outputs. Users should avoid over-reliance in high-stakes domains (e.g., healthcare, law, and finance). Second, Stable-RAG relies on external documents for grounding, which may contain biases or errors, potentially propagating or amplifying them.

\section*{Acknowledgements}
This work was funded by the National Natural Science Foundation of China (NSFC) under Grants No. U25B2070 and 62406013, the Beijing Advanced Innovation Center Funds for Future Blockchain and Privacy Computing (GJJ-24-034), and the Fundamental Research Funds for the Central Universities.
\bibliography{custom}

@article{nq,
    title = "Natural Questions: A Benchmark for Question Answering Research",
    author = "Kwiatkowski, Tom  and
      Palomaki, Jennimaria  and
      Redfield, Olivia  and
      Collins, Michael  and
      Parikh, Ankur  and
      Alberti, Chris  and
      Epstein, Danielle  and
      Polosukhin, Illia  and
      Devlin, Jacob  and
      Lee, Kenton  and
      Toutanova, Kristina  and
      Jones, Llion  and
      Kelcey, Matthew  and
      Chang, Ming-Wei  and
      Dai, Andrew M.  and
      Uszkoreit, Jakob  and
      Le, Quoc  and
      Petrov, Slav",
    journal = "Transactions of the Association for Computational Linguistics",
    volume = "7",
    year = "2019",
    address = "Cambridge, MA",
    publisher = "MIT Press",
    url = "https://aclanthology.org/Q19-1026/",
    doi = "10.1162/tacl_a_00276",
    pages = "452--466",
}

@inproceedings{triviaqa,
    title = "{T}rivia{QA}: A Large Scale Distantly Supervised Challenge Dataset for Reading Comprehension",
    author = "Joshi, Mandar  and
      Choi, Eunsol  and
      Weld, Daniel  and
      Zettlemoyer, Luke",
    booktitle = "Proceedings of the 55th Annual Meeting of the Association for Computational Linguistics (Volume 1: Long Papers)",
    year = "2017",
    url = "https://aclanthology.org/P17-1147/",
    doi = "10.18653/v1/P17-1147",
    pages = "1601--1611",
}

@inproceedings{hotpotqa,
    title = "{H}otpot{QA}: A Dataset for Diverse, Explainable Multi-hop Question Answering",
    author = "Yang, Zhilin  and
      Qi, Peng  and
      Zhang, Saizheng  and
      Bengio, Yoshua  and
      Cohen, William  and
      Salakhutdinov, Ruslan  and
      Manning, Christopher D.",
    booktitle = "EMNLP",
    month = oct # "-" # nov,
    year = "2018",
    url = "https://aclanthology.org/D18-1259",
    doi = "10.18653/v1/D18-1259",
    pages = "2369--2380",
}

@inproceedings{zhu-etal-2024-atm,
    title = "{ATM}: Adversarial Tuning Multi-agent System Makes a Robust Retrieval-Augmented Generator",
    author = "Zhu, Junda  and
      Yan, Lingyong  and
      Shi, Haibo  and
      Yin, Dawei  and
      Sha, Lei",
    booktitle = "Proceedings of the 2024 Conference on Empirical Methods in Natural Language Processing",
    month = nov,
    year = "2024",
    url = "https://aclanthology.org/2024.emnlp-main.610/",
    doi = "10.18653/v1/2024.emnlp-main.610",
    pages = "10902--10919",

}

@inproceedings{dpr,
    title = "Dense Passage Retrieval for Open-Domain Question Answering",
    author = "Karpukhin, Vladimir  and
      Oguz, Barlas  and
      Min, Sewon  and
      Lewis, Patrick  and
      Wu, Ledell  and
      Edunov, Sergey  and
      Chen, Danqi  and
      Yih, Wen-tau",
    editor = "Webber, Bonnie  and
      Cohn, Trevor  and
      He, Yulan  and
      Liu, Yang",
    booktitle = "Proceedings of the 2020 Conference on Empirical Methods in Natural Language Processing (EMNLP)",
    month = nov,
    year = "2020",
    address = "Online",
    publisher = "Association for Computational Linguistics",
    url = "https://aclanthology.org/2020.emnlp-main.550/",
    doi = "10.18653/v1/2020.emnlp-main.550",
    pages = "6769--6781"
}

@inproceedings{retrobust,
  title={Making Retrieval-Augmented Language Models Robust to Irrelevant Context},
  author={Yoran, Ori and Wolfson, Tomer and Ram, Ori and Berant, Jonathan},
  booktitle={The Twelfth International Conference on Learning Representations},
  year={2024}
}

@inproceedings{raat,
    title = "Enhancing Noise Robustness of Retrieval-Augmented Language Models with Adaptive Adversarial Training",
    author = "Fang, Feiteng  and
      Bai, Yuelin  and
      Ni, Shiwen  and
      Yang, Min  and
      Chen, Xiaojun  and
      Xu, Ruifeng",
    booktitle = "Proceedings of the 62nd Annual Meeting of the Association for Computational Linguistics (Volume 1: Long Papers)",
    year = "2024",
    url = "https://aclanthology.org/2024.acl-long.540/",
    doi = "10.18653/v1/2024.acl-long.540",
    pages = "10028--10039",
}

@article{mspoe,
  title={Found in the middle: How language models use long contexts better via plug-and-play positional encoding},
  author={Zhang, Zhenyu and Chen, Runjin and Liu, Shiwei and Yao, Zhewei and Ruwase, Olatunji and Chen, Beidi and Wu, Xiaoxia and Wang, Zhangyang},
  journal={Advances in Neural Information Processing Systems},
  volume={37},
  pages={60755--60775},
  year={2024}
}

@inproceedings{ni-etal-2025-towards,
    title = "Towards Fully Exploiting {LLM} Internal States to Enhance Knowledge Boundary Perception",
    author = "Ni, Shiyu  and
      Bi, Keping  and
      Guo, Jiafeng  and
      Yu, Lulu  and
      Bi, Baolong  and
      Cheng, Xueqi",
    booktitle = "Proceedings of the 63rd Annual Meeting of the Association for Computational Linguistics (Volume 1: Long Papers)",
    year = "2025",
    url = "https://aclanthology.org/2025.acl-long.1184/",
    doi = "10.18653/v1/2025.acl-long.1184",
    pages = "24315--24329",
}

@article{izacard2021unsupervised,
  title={Unsupervised dense information retrieval with contrastive learning},
  author={Izacard, Gautier and Caron, Mathilde and Hosseini, Lucas and Riedel, Sebastian and Bojanowski, Piotr and Joulin, Armand and Grave, Edouard},
  journal={arXiv preprint arXiv:2112.09118},
  year={2021}
}

@inproceedings{wang-etal-2025-position,
    title = "{POSITION} {BIAS} {MITIGATES} {POSITION} {BIAS}: Mitigate Position Bias Through Inter-Position Knowledge Distillation",
    author = "Wang, Yifei  and
      Xiong, Feng  and
      Wang, Yong  and
      Li, Linjing  and
      Chu, Xiangxiang  and
      Zeng, Daniel Dajun",
    booktitle = "Proceedings of the 2025 Conference on Empirical Methods in Natural Language Processing",
    year = "2025",
    url = "https://aclanthology.org/2025.emnlp-main.78/",
    doi = "10.18653/v1/2025.emnlp-main.78",
    pages = "1495--1512",
}

@misc{llama3,
      title={The Llama 3 Herd of Models}, 
      author={Aaron Grattafiori and Abhimanyu Dubey and Abhinav Jauhri and Abhinav Pandey and Abhishek Kadian and Ahmad Al-Dahle and Aiesha Letman and Akhil Mathur and Alan Schelten and Alex Vaughan and Amy Yang and Angela Fan and Anirudh Goyal and Anthony Hartshorn and Aobo Yang and Archi Mitra and Archie Sravankumar and Artem Korenev and Arthur Hinsvark and Arun Rao and Aston Zhang and Aurelien Rodriguez and Austen Gregerson and Ava Spataru and Baptiste Roziere and Bethany Biron and Binh Tang and Bobbie Chern and Charlotte Caucheteux and Chaya Nayak and Chloe Bi and Chris Marra and Chris McConnell and Christian Keller and Christophe Touret and Chunyang Wu and Corinne Wong and Cristian Canton Ferrer and Cyrus Nikolaidis and Damien Allonsius and Daniel Song and Danielle Pintz and Danny Livshits and Danny Wyatt and David Esiobu and Dhruv Choudhary and Dhruv Mahajan and Diego Garcia-Olano and Diego Perino and Dieuwke Hupkes and Egor Lakomkin and Ehab AlBadawy and Elina Lobanova and Emily Dinan and Eric Michael Smith and Filip Radenovic and Francisco Guzmán and Frank Zhang and Gabriel Synnaeve and Gabrielle Lee and Georgia Lewis Anderson and Govind Thattai and Graeme Nail and Gregoire Mialon and Guan Pang and Guillem Cucurell and Hailey Nguyen and Hannah Korevaar and Hu Xu and Hugo Touvron and Iliyan Zarov and Imanol Arrieta Ibarra and Isabel Kloumann and Ishan Misra and Ivan Evtimov and Jack Zhang and Jade Copet and Jaewon Lee and Jan Geffert and Jana Vranes and Jason Park and Jay Mahadeokar and Jeet Shah and Jelmer van der Linde and Jennifer Billock and Jenny Hong and Jenya Lee and Jeremy Fu and Jianfeng Chi and Jianyu Huang and Jiawen Liu and Jie Wang and Jiecao Yu and Joanna Bitton and Joe Spisak and Jongsoo Park and Joseph Rocca and Joshua Johnstun and Joshua Saxe and Junteng Jia and Kalyan Vasuden Alwala and Karthik Prasad and Kartikeya Upasani and Kate Plawiak and Ke Li and Kenneth Heafield and Kevin Stone and Khalid El-Arini and Krithika Iyer and Kshitiz Malik and Kuenley Chiu and Kunal Bhalla and Kushal Lakhotia and Lauren Rantala-Yeary and Laurens van der Maaten and Lawrence Chen and Liang Tan and Liz Jenkins and Louis Martin and Lovish Madaan and Lubo Malo and Lukas Blecher and Lukas Landzaat and Luke de Oliveira and Madeline Muzzi and Mahesh Pasupuleti and Mannat Singh and Manohar Paluri and Marcin Kardas and Maria Tsimpoukelli and Mathew Oldham and Mathieu Rita and Maya Pavlova and Melanie Kambadur and Mike Lewis and Min Si and Mitesh Kumar Singh and Mona Hassan and Naman Goyal and Narjes Torabi and Nikolay Bashlykov and Nikolay Bogoychev and Niladri Chatterji and Ning Zhang and Olivier Duchenne and Onur Çelebi and Patrick Alrassy and Pengchuan Zhang and Pengwei Li and Petar Vasic and Peter Weng and Prajjwal Bhargava and Pratik Dubal and Praveen Krishnan and Punit Singh Koura and Puxin Xu and Qing He and Qingxiao Dong and Ragavan Srinivasan and Raj Ganapathy and Ramon Calderer and Ricardo Silveira Cabral and Robert Stojnic and Roberta Raileanu and Rohan Maheswari and Rohit Girdhar and Rohit Patel and Romain Sauvestre and Ronnie Polidoro and Roshan Sumbaly and Ross Taylor and Ruan Silva and Rui Hou and Rui Wang and Saghar Hosseini and Sahana Chennabasappa and Sanjay Singh and Sean Bell and Seohyun Sonia Kim and Sergey Edunov and Shaoliang Nie and Sharan Narang and Sharath Raparthy and Sheng Shen and Shengye Wan and Shruti Bhosale and Shun Zhang and Simon Vandenhende and Soumya Batra and Spencer Whitman and Sten Sootla and Stephane Collot and Suchin Gururangan and Sydney Borodinsky and Tamar Herman and Tara Fowler and Tarek Sheasha and Thomas Georgiou and Thomas Scialom and Tobias Speckbacher and Todor Mihaylov and Tong Xiao and Ujjwal Karn and Vedanuj Goswami and Vibhor Gupta and Vignesh Ramanathan and Viktor Kerkez and Vincent Gonguet and Virginie Do and Vish Vogeti and Vítor Albiero and Vladan Petrovic and Weiwei Chu and Wenhan Xiong and Wenyin Fu and Whitney Meers and Xavier Martinet and Xiaodong Wang and Xiaofang Wang and Xiaoqing Ellen Tan and Xide Xia and Xinfeng Xie and Xuchao Jia and Xuewei Wang and Yaelle Goldschlag and Yashesh Gaur and Yasmine Babaei and Yi Wen and Yiwen Song and Yuchen Zhang and Yue Li and Yuning Mao and Zacharie Delpierre Coudert and Zheng Yan and Zhengxing Chen and Zoe Papakipos and Aaditya Singh and Aayushi Srivastava and Abha Jain and Adam Kelsey and Adam Shajnfeld and Adithya Gangidi and Adolfo Victoria and Ahuva Goldstand and Ajay Menon and Ajay Sharma and Alex Boesenberg and Alexei Baevski and Allie Feinstein and Amanda Kallet and Amit Sangani and Amos Teo and Anam Yunus and Andrei Lupu and Andres Alvarado and Andrew Caples and Andrew Gu and Andrew Ho and Andrew Poulton and Andrew Ryan and Ankit Ramchandani and Annie Dong and Annie Franco and Anuj Goyal and Aparajita Saraf and Arkabandhu Chowdhury and Ashley Gabriel and Ashwin Bharambe and Assaf Eisenman and Azadeh Yazdan and Beau James and Ben Maurer and Benjamin Leonhardi and Bernie Huang and Beth Loyd and Beto De Paola and Bhargavi Paranjape and Bing Liu and Bo Wu and Boyu Ni and Braden Hancock and Bram Wasti and Brandon Spence and Brani Stojkovic and Brian Gamido and Britt Montalvo and Carl Parker and Carly Burton and Catalina Mejia and Ce Liu and Changhan Wang and Changkyu Kim and Chao Zhou and Chester Hu and Ching-Hsiang Chu and Chris Cai and Chris Tindal and Christoph Feichtenhofer and Cynthia Gao and Damon Civin and Dana Beaty and Daniel Kreymer and Daniel Li and David Adkins and David Xu and Davide Testuggine and Delia David and Devi Parikh and Diana Liskovich and Didem Foss and Dingkang Wang and Duc Le and Dustin Holland and Edward Dowling and Eissa Jamil and Elaine Montgomery and Eleonora Presani and Emily Hahn and Emily Wood and Eric-Tuan Le and Erik Brinkman and Esteban Arcaute and Evan Dunbar and Evan Smothers and Fei Sun and Felix Kreuk and Feng Tian and Filippos Kokkinos and Firat Ozgenel and Francesco Caggioni and Frank Kanayet and Frank Seide and Gabriela Medina Florez and Gabriella Schwarz and Gada Badeer and Georgia Swee and Gil Halpern and Grant Herman and Grigory Sizov and Guangyi and Zhang and Guna Lakshminarayanan and Hakan Inan and Hamid Shojanazeri and Han Zou and Hannah Wang and Hanwen Zha and Haroun Habeeb and Harrison Rudolph and Helen Suk and Henry Aspegren and Hunter Goldman and Hongyuan Zhan and Ibrahim Damlaj and Igor Molybog and Igor Tufanov and Ilias Leontiadis and Irina-Elena Veliche and Itai Gat and Jake Weissman and James Geboski and James Kohli and Janice Lam and Japhet Asher and Jean-Baptiste Gaya and Jeff Marcus and Jeff Tang and Jennifer Chan and Jenny Zhen and Jeremy Reizenstein and Jeremy Teboul and Jessica Zhong and Jian Jin and Jingyi Yang and Joe Cummings and Jon Carvill and Jon Shepard and Jonathan McPhie and Jonathan Torres and Josh Ginsburg and Junjie Wang and Kai Wu and Kam Hou U and Karan Saxena and Kartikay Khandelwal and Katayoun Zand and Kathy Matosich and Kaushik Veeraraghavan and Kelly Michelena and Keqian Li and Kiran Jagadeesh and Kun Huang and Kunal Chawla and Kyle Huang and Lailin Chen and Lakshya Garg and Lavender A and Leandro Silva and Lee Bell and Lei Zhang and Liangpeng Guo and Licheng Yu and Liron Moshkovich and Luca Wehrstedt and Madian Khabsa and Manav Avalani and Manish Bhatt and Martynas Mankus and Matan Hasson and Matthew Lennie and Matthias Reso and Maxim Groshev and Maxim Naumov and Maya Lathi and Meghan Keneally and Miao Liu and Michael L. Seltzer and Michal Valko and Michelle Restrepo and Mihir Patel and Mik Vyatskov and Mikayel Samvelyan and Mike Clark and Mike Macey and Mike Wang and Miquel Jubert Hermoso and Mo Metanat and Mohammad Rastegari and Munish Bansal and Nandhini Santhanam and Natascha Parks and Natasha White and Navyata Bawa and Nayan Singhal and Nick Egebo and Nicolas Usunier and Nikhil Mehta and Nikolay Pavlovich Laptev and Ning Dong and Norman Cheng and Oleg Chernoguz and Olivia Hart and Omkar Salpekar and Ozlem Kalinli and Parkin Kent and Parth Parekh and Paul Saab and Pavan Balaji and Pedro Rittner and Philip Bontrager and Pierre Roux and Piotr Dollar and Polina Zvyagina and Prashant Ratanchandani and Pritish Yuvraj and Qian Liang and Rachad Alao and Rachel Rodriguez and Rafi Ayub and Raghotham Murthy and Raghu Nayani and Rahul Mitra and Rangaprabhu Parthasarathy and Raymond Li and Rebekkah Hogan and Robin Battey and Rocky Wang and Russ Howes and Ruty Rinott and Sachin Mehta and Sachin Siby and Sai Jayesh Bondu and Samyak Datta and Sara Chugh and Sara Hunt and Sargun Dhillon and Sasha Sidorov and Satadru Pan and Saurabh Mahajan and Saurabh Verma and Seiji Yamamoto and Sharadh Ramaswamy and Shaun Lindsay and Shaun Lindsay and Sheng Feng and Shenghao Lin and Shengxin Cindy Zha and Shishir Patil and Shiva Shankar and Shuqiang Zhang and Shuqiang Zhang and Sinong Wang and Sneha Agarwal and Soji Sajuyigbe and Soumith Chintala and Stephanie Max and Stephen Chen and Steve Kehoe and Steve Satterfield and Sudarshan Govindaprasad and Sumit Gupta and Summer Deng and Sungmin Cho and Sunny Virk and Suraj Subramanian and Sy Choudhury and Sydney Goldman and Tal Remez and Tamar Glaser and Tamara Best and Thilo Koehler and Thomas Robinson and Tianhe Li and Tianjun Zhang and Tim Matthews and Timothy Chou and Tzook Shaked and Varun Vontimitta and Victoria Ajayi and Victoria Montanez and Vijai Mohan and Vinay Satish Kumar and Vishal Mangla and Vlad Ionescu and Vlad Poenaru and Vlad Tiberiu Mihailescu and Vladimir Ivanov and Wei Li and Wenchen Wang and Wenwen Jiang and Wes Bouaziz and Will Constable and Xiaocheng Tang and Xiaojian Wu and Xiaolan Wang and Xilun Wu and Xinbo Gao and Yaniv Kleinman and Yanjun Chen and Ye Hu and Ye Jia and Ye Qi and Yenda Li and Yilin Zhang and Ying Zhang and Yossi Adi and Youngjin Nam and Yu and Wang and Yu Zhao and Yuchen Hao and Yundi Qian and Yunlu Li and Yuzi He and Zach Rait and Zachary DeVito and Zef Rosnbrick and Zhaoduo Wen and Zhenyu Yang and Zhiwei Zhao and Zhiyu Ma},
      year={2024},
      eprint={2407.21783},
      archivePrefix={arXiv},
      primaryClass={cs.AI},
      url={https://arxiv.org/abs/2407.21783}, 
}

@article{qwen3,
  title={Qwen3 technical report},
  author={An Yang and Anfeng Li and Baosong Yang and Beichen Zhang and Binyuan Hui and Bo Zheng and Bowen Yu and Chang Gao and Chengen Huang and Chenxu Lv and Chujie Zheng and Dayiheng Liu and Fan Zhou and Fei Huang and Feng Hu and Hao Ge and Haoran Wei and Huan Lin and Jialong Tang and Jian Yang and Jianhong Tu and Jianwei Zhang and Jianxin Yang and Jiaxi Yang and Jing Zhou and Jingren Zhou and Junyang Lin and Kai Dang and Keqin Bao and Kexin Yang and Le Yu and Lianghao Deng and Mei Li and Mingfeng Xue and Mingze Li and Pei Zhang and Peng Wang and Qin Zhu and Rui Men and Ruize Gao and Shixuan Liu and Shuang Luo and Tianhao Li and Tianyi Tang and Wenbiao Yin and Xingzhang Ren and Xinyu Wang and Xinyu Zhang and Xuancheng Ren and Yang Fan and Yang Su and Yichang Zhang and Yinger Zhang and Yu Wan and Yuqiong Liu and Zekun Wang and Zeyu Cui and Zhenru Zhang and Zhipeng Zhou and Zihan Qiu},
  journal={arXiv preprint arXiv:2505.09388},
  year={2025}
}

@inproceedings{cuconasu2024power,
  title={The power of noise: Redefining retrieval for rag systems},
  author={Cuconasu, Florin and Trappolini, Giovanni and Siciliano, Federico and Filice, Simone and Campagnano, Cesare and Maarek, Yoelle and Tonellotto, Nicola and Silvestri, Fabrizio},
  booktitle={Proceedings of the 47th International ACM SIGIR Conference on Research and Development in Information Retrieval},
  pages={719--729},
  year={2024}
}

@inproceedings{xu2024recomp,
  title={RECOMP: Improving retrieval-augmented LMs with context compression and selective augmentation},
  author={Xu, Fangyuan and Shi, Weijia and Choi, Eunsol},
  booktitle={The Twelfth International Conference on Learning Representations},
  year={2024}
}

@inproceedings{compselect,
author = {Zhang, Qianchi and Zhang, Hainan and Pang, Liang and Tong, Yongxin and Zheng, Hongwei and Zheng, Zhiming},
title = {Less is More: Compact Clue Selection for Efficient Retrieval-Augmented Generation Reasoning},
year = {2026},
url = {https://doi.org/10.1145/3774904.3792158},
doi = {10.1145/3774904.3792158},
booktitle = {Proceedings of the ACM Web Conference 2026},
pages = {1971--1982},
series = {WWW '26}
}

@inproceedings{lewis2020retrieval,
 author = {Lewis, Patrick and Perez, Ethan and Piktus, Aleksandra and Petroni, Fabio and Karpukhin, Vladimir and Goyal, Naman and K\"{u}ttler, Heinrich and Lewis, Mike and Yih, Wen-tau and Rockt\"{a}schel, Tim and Riedel, Sebastian and Kiela, Douwe},
 booktitle = {Advances in Neural Information Processing Systems},
 editor = {H. Larochelle and M. Ranzato and R. Hadsell and M.F. Balcan and H. Lin},
 pages = {9459--9474},
 publisher = {Curran Associates, Inc.},
 title = {Retrieval-Augmented Generation for Knowledge-Intensive NLP Tasks},
 url = {https://proceedings.neurips.cc/paper_files/paper/2020/file/6b493230205f780e1bc26945df7481e5-Paper.pdf},
 volume = {33},
 year = {2020}
}

@inproceedings{fan2024survey,
  title={A survey on rag meeting llms: Towards retrieval-augmented large language models},
  author={Fan, Wenqi and Ding, Yujuan and Ning, Liangbo and Wang, Shijie and Li, Hengyun and Yin, Dawei and Chua, Tat-Seng and Li, Qing},
  booktitle={Proceedings of the 30th ACM SIGKDD conference on knowledge discovery and data mining},
  pages={6491--6501},
  year={2024}
}

@inproceedings{chen2022gere,
  title={GERE: Generative evidence retrieval for fact verification},
  author={Chen, Jiangui and Zhang, Ruqing and Guo, Jiafeng and Fan, Yixing and Cheng, Xueqi},
  booktitle={Proceedings of the 45th International ACM SIGIR Conference on Research and Development in Information Retrieval},
  pages={2184--2189},
  year={2022}
}

@inproceedings{huang-etal-2023-learning,
    title = "Learning Retrieval Augmentation for Personalized Dialogue Generation",
    author = "Huang, Qiushi  and
      Fu, Shuai  and
      Liu, Xubo  and
      Wang, Wenwu  and
      Ko, Tom  and
      Zhang, Yu  and
      Tang, Lilian",
    booktitle = "Proceedings of the 2023 Conference on Empirical Methods in Natural Language Processing",
    year = "2023",
    url = "https://aclanthology.org/2023.emnlp-main.154/",
    doi = "10.18653/v1/2023.emnlp-main.154",
    pages = "2523--2540",
}

@article{peysakhovich2023attention,
  title={Attention sorting combats recency bias in long context language models},
  author={Peysakhovich, Alexander and Lerer, Adam},
  journal={arXiv preprint arXiv:2310.01427},
  year={2023}
}

@article{su2024roformer,
  title={Roformer: Enhanced transformer with rotary position embedding},
  author={Su, Jianlin and Ahmed, Murtadha and Lu, Yu and Pan, Shengfeng and Bo, Wen and Liu, Yunfeng},
  journal={Neurocomputing},
  volume={568},
  pages={127063},
  year={2024},
  publisher={Elsevier}
}

@article{press2021train,
  title={Train short, test long: Attention with linear biases enables input length extrapolation},
  author={Press, Ofir and Smith, Noah A and Lewis, Mike},
  journal={arXiv preprint arXiv:2108.12409},
  year={2021}
}

@inproceedings{xiaoefficient,
  title={Efficient Streaming Language Models with Attention Sinks},
  author={Xiao, Guangxuan and Tian, Yuandong and Chen, Beidi and Han, Song and Lewis, Mike},
  booktitle={The Twelfth International Conference on Learning Representations}
}

@inproceedings{guattention,
  title={When Attention Sink Emerges in Language Models: An Empirical View},
  author={Gu, Xiangming and Pang, Tianyu and Du, Chao and Liu, Qian and Zhang, Fengzhuo and Du, Cunxiao and Wang, Ye and Lin, Min},
  booktitle={The Thirteenth International Conference on Learning Representations}
}

@inproceedings{chen-etal-2024-fortify,
    title = "Fortify the Shortest Stave in Attention: Enhancing Context Awareness of Large Language Models for Effective Tool Use",
    author = "Chen, Yuhan  and
      Lv, Ang  and
      Lin, Ting-En  and
      Chen, Changyu  and
      Wu, Yuchuan  and
      Huang, Fei  and
      Li, Yongbin  and
      Yan, Rui",
    booktitle = "Proceedings of the 62nd Annual Meeting of the Association for Computational Linguistics (Volume 1: Long Papers)",
    year = "2024",
    url = "https://aclanthology.org/2024.acl-long.601/",
    doi = "10.18653/v1/2024.acl-long.601",
    pages = "11160--11174",
    
}

@article{lin2024mixture,
  title={Mixture of in-context experts enhance llms' long context awareness},
  author={Lin, Hongzhan and Lv, Ang and Chen, Yuhan and Zhu, Chen and Song, Yang and Zhu, Hengshu and Yan, Rui},
  journal={Advances in Neural Information Processing Systems},
  volume={37},
  pages={79573--79596},
  year={2024}
}

@inproceedings{hsieh-etal-2024-found,
    title = "Found in the middle: Calibrating Positional Attention Bias Improves Long Context Utilization",
    author = "Hsieh, Cheng-Yu  and
      Chuang, Yung-Sung  and
      Li, Chun-Liang  and
      Wang, Zifeng  and
      Le, Long  and
      Kumar, Abhishek  and
      Glass, James  and
      Ratner, Alexander  and
      Lee, Chen-Yu  and
      Krishna, Ranjay  and
      Pfister, Tomas",
    booktitle = "Findings of the Association for Computational Linguistics: ACL 2024",
    year = "2024",
    url = "https://aclanthology.org/2024.findings-acl.890/",
    doi = "10.18653/v1/2024.findings-acl.890",
    pages = "14982--14995",
}

@misc{liang2025clue,
      title={CLUE: Non-parametric Verification from Experience via Hidden-State Clustering}, 
      author={Zhenwen Liang and Ruosen Li and Yujun Zhou and Linfeng Song and Dian Yu and Xinya Du and Haitao Mi and Dong Yu},
      year={2025},
      eprint={2510.01591},
      archivePrefix={arXiv},
      primaryClass={cs.CL},
      url={https://arxiv.org/abs/2510.01591}, 
}

@inproceedings{lee-etal-2025-efficient-latent,
    title = "Efficient Latent Semantic Clustering for Scaling Test-Time Computation of {LLM}s",
    author = "Lee, Sungjae  and
      Kim, Hoyoung  and
      Hwang, Jeongyeon  and
      Park, Eunhyeok  and
      Ok, Jungseul",
    booktitle = "Findings of the Association for Computational Linguistics: EMNLP 2025",
    year = "2025",
    url = "https://aclanthology.org/2025.findings-emnlp.1310/",
    doi = "10.18653/v1/2025.findings-emnlp.1310",
    pages = "24126--24144",

}

@inproceedings{zhang-etal-2024-r,
    title = "{R}-Tuning: Instructing Large Language Models to Say `{I} Don{'}t Know'",
    author = "Zhang, Hanning  and
      Diao, Shizhe  and
      Lin, Yong  and
      Fung, Yi  and
      Lian, Qing  and
      Wang, Xingyao  and
      Chen, Yangyi  and
      Ji, Heng  and
      Zhang, Tong",
    booktitle = "Proceedings of the 2024 Conference of the North American Chapter of the Association for Computational Linguistics: Human Language Technologies (Volume 1: Long Papers)",
    year = "2024",
    url = "https://aclanthology.org/2024.naacl-long.394/",
    doi = "10.18653/v1/2024.naacl-long.394",
    pages = "7113--7139",
}

@article{rafailov2023direct,
  title={Direct preference optimization: Your language model is secretly a reward model},
  author={Rafailov, Rafael and Sharma, Archit and Mitchell, Eric and Manning, Christopher D and Ermon, Stefano and Finn, Chelsea},
  journal={Advances in neural information processing systems},
  volume={36},
  pages={53728--53741},
  year={2023}
}

@article{zhang2024adacomp,
  title={Adacomp: Extractive context compression with adaptive predictor for retrieval-augmented large language models},
  author={Zhang, Qianchi and Zhang, Hainan and Pang, Liang and Zheng, Hongwei and Zheng, Zhiming},
  journal={arXiv preprint arXiv:2409.01579},
  year={2024}
}

@misc{gao2023retrieval,
      title={Retrieval-Augmented Generation for Large Language Models: A Survey}, 
      author={Yunfan Gao and Yun Xiong and Xinyu Gao and Kangxiang Jia and Jinliu Pan and Yuxi Bi and Yi Dai and Jiawei Sun and Meng Wang and Haofen Wang},
      year={2024},
      eprint={2312.10997},
      archivePrefix={arXiv},
      primaryClass={cs.CL},
      url={https://arxiv.org/abs/2312.10997}, 
}

@article{zhou2024trustworthiness,
  title={Trustworthiness in retrieval-augmented generation systems: A survey},
  author={Zhou, Yujia and Liu, Yan and Li, Xiaoxi and Jin, Jiajie and Qian, Hongjin and Liu, Zheng and Li, Chaozhuo and Dou, Zhicheng and Ho, Tsung-Yi and Yu, Philip S},
  journal={arXiv preprint arXiv:2409.10102},
  year={2024}
}

@inproceedings{wolf-etal-2020-transformers,
    title = "Transformers: State-of-the-Art Natural Language Processing",
    author = "Wolf, Thomas  and
      Debut, Lysandre  and
      Sanh, Victor  and
      Chaumond, Julien  and
      Delangue, Clement  and
      Moi, Anthony  and
      Cistac, Pierric  and
      Rault, Tim  and
      Louf, Remi  and
      Funtowicz, Morgan  and
      Davison, Joe  and
      Shleifer, Sam  and
      von Platen, Patrick  and
      Ma, Clara  and
      Jernite, Yacine  and
      Plu, Julien  and
      Xu, Canwen  and
      Le Scao, Teven  and
      Gugger, Sylvain  and
      Drame, Mariama  and
      Lhoest, Quentin  and
      Rush, Alexander",
    booktitle = "Proceedings of the 2020 Conference on Empirical Methods in Natural Language Processing: System Demonstrations",
    year = "2020",
    url = "https://aclanthology.org/2020.emnlp-demos.6/",
    doi = "10.18653/v1/2020.emnlp-demos.6",
    pages = "38--45",
   
}

@inproceedings{
hu2022lora,
title={Lo{RA}: Low-Rank Adaptation of Large Language Models},
author={Edward J Hu and Yelong Shen and Phillip Wallis and Zeyuan Allen-Zhu and Yuanzhi Li and Shean Wang and Lu Wang and Weizhu Chen},
booktitle={International Conference on Learning Representations},
year={2022},
url={https://openreview.net/forum?id=nZeVKeeFYf9}
}

@article{von2007tutorial,
  title={A tutorial on spectral clustering},
  author={Von Luxburg, Ulrike},
  journal={Statistics and computing},
  volume={17},
  number={4},
  pages={395--416},
  year={2007},
  publisher={Springer}
}

@article{ng2001spectral,
  title={On spectral clustering: Analysis and an algorithm},
  author={Ng, Andrew and Jordan, Michael and Weiss, Yair},
  journal={Advances in neural information processing systems},
  volume={14},
  year={2001}
}

@inproceedings{azaria-mitchell-2023-internal,
    title = "The Internal State of an {LLM} Knows When It{'}s Lying",
    author = "Azaria, Amos  and
      Mitchell, Tom",
    booktitle = "Findings of the Association for Computational Linguistics: EMNLP 2023",
    year = "2023",
    url = "https://aclanthology.org/2023.findings-emnlp.68/",
    doi = "10.18653/v1/2023.findings-emnlp.68",
    pages = "967--976",
}

@article{chen2025privacy,
  title={Privacy-Preserving Reasoning with Knowledge-Distilled Parametric Retrieval Augmented Generation},
  author={Chen, Jinwen and Zhang, Hainan and Pang, Liang and Tong, Yongxin and Zhou, Haibo and Zhan, Yuan and Lin, Wei and Zheng, Zhiming},
  journal={arXiv preprint arXiv:2509.01088},
  year={2025}
}

@article{wang2026fbs,
  title={FBS: Modeling Native Parallel Reading inside a Transformer},
  author={Wang, Tongxi},
  journal={arXiv preprint arXiv:2601.21708},
  year={2026}
}

@inproceedings{luo-etal-2025-dtcrs,
    title = "{DTCRS}: Dynamic Tree Construction for Recursive Summarization",
    author = "Luo, Guanran  and
      Jian, Zhongquan  and
      Qiu, Wentao  and
      Wang, Meihong  and
      Wu, Qingqiang",
    booktitle = "Proceedings of the 63rd Annual Meeting of the Association for Computational Linguistics (Volume 1: Long Papers)",
    year = "2025",
    url = "https://aclanthology.org/2025.acl-long.536/",
    doi = "10.18653/v1/2025.acl-long.536",
    pages = "10948--10963",
}

@misc{luo2026agscadaptivegranularitysemantic,
      title={AGSC: Adaptive Granularity and Semantic Clustering for Uncertainty Quantification in Long-text Generation}, 
      author={Guanran Luo and Wentao Qiu and Wanru Zhao and Wenhan Lv and Zhongquan Jian and Meihong Wang and Qingqiang Wu},
      year={2026},
      eprint={2604.06812},
      archivePrefix={arXiv},
      primaryClass={cs.CL},
      url={https://arxiv.org/abs/2604.06812}, 
}

@misc{luo2026gcotdecodingunlockingdeepreasoning,
      title={GCoT-Decoding: Unlocking Deep Reasoning Paths for Universal Question Answering}, 
      author={Guanran Luo and Wentao Qiu and Zhongquan Jian and Meihong Wang and Qingqiang Wu},
      year={2026},
      eprint={2604.06794},
      archivePrefix={arXiv},
      primaryClass={cs.CL},
      url={https://arxiv.org/abs/2604.06794}, 
}

@inproceedings{wang2025maferw,
  title={Maferw: Query rewriting with multi-aspect feedbacks for retrieval-augmented large language models},
  author={Wang, Yujing and Zhang, Hainan and Pang, Liang and Guo, Binghui and Zheng, Hongwei and Zheng, Zhiming},
  booktitle={Proceedings of the AAAI Conference on Artificial Intelligence},
  volume={39},
  number={24},
  pages={25434--25442},
  year={2025}
}

@misc{ji2026strideedstrategygroundedstepwisereasoning,
      title={STRIDE-ED: A Strategy-Grounded Stepwise Reasoning Framework for Empathetic Dialogue Systems}, 
      author={Hongru Ji and Yuyin Fan and Meng Zhao and Xianghua Li and Lianwei Wu and Chao Gao},
      year={2026},
      eprint={2604.07100},
      archivePrefix={arXiv},
      primaryClass={cs.CL},
      url={https://arxiv.org/abs/2604.07100}, 
}

@inproceedings{li-etal-2025-misleading,
    title = "From Misleading Queries to Accurate Answers: A Three-Stage Fine-Tuning Method for {LLM}s",
    author = "Li, Guocong  and
      Liu, Weize  and
      Wu, Yihang  and
      Wang, Ping  and
      Huang, Shuaihan  and
      Xu, Hongxia  and
      Wu, Jian",
    booktitle = "Findings of the Association for Computational Linguistics: ACL 2025",
    year = "2025",
    url = "https://aclanthology.org/2025.findings-acl.65/",
    doi = "10.18653/v1/2025.findings-acl.65",
    pages = "1192--1209",
}

@inproceedings{
limol,
title={MoL: Adaptive Mixture-of-Length Reasoning for Efficient Question Answering with Context},
author={Guocong Li and Jinjian Zhang and Ping Wang and Dongnan Liu and Tian Liang and Qiuyi Qi and Hao Huang and Siyan Guo and Mutian Bao and Wei Zhou and Linjian Mo and Hongxia Xu and Jian Wu},
booktitle={The Fourteenth International Conference on Learning Representations},
year={2026},
url={https://openreview.net/forum?id=oWWAeLEdE3}
}

@inproceedings{ma-etal-2024-context,
    title = "Context-Driven Index Trimming: A Data Quality Perspective to Enhancing Precision of {RALM}s",
    author = "Ma, Kexin  and
      Jin, Ruochun  and
      Haotian, Wang  and
      Xi, Wang  and
      Chen, Huan  and
      Tang, Yuhua  and
      Wang, Qian",
    booktitle = "Findings of the Association for Computational Linguistics: EMNLP 2024",
    year = "2024",
    url = "https://aclanthology.org/2024.findings-emnlp.281/",
    doi = "10.18653/v1/2024.findings-emnlp.281",
    pages = "4886--4901",
}

@misc{ma2026cast,
      title={CAST: Character-and-Scene Episodic Memory for Agents}, 
      author={Kexin Ma and Bojun Li and Yuhua Tang and Liting Sun and Ruochun Jin},
      year={2026},
      eprint={2602.06051},
      archivePrefix={arXiv},
      primaryClass={cs.CL},
      url={https://arxiv.org/abs/2602.06051}, 
}

@article{hu2026contextagent,
  title={Context-Agent: Dynamic Discourse Trees for Non-Linear Dialogue},
  author={Hu, Junan and Guo, Shudan and Liu, Wenqi and Yin, Jianhua and Wei, Yinwei},
  journal={arXiv preprint arXiv:2604.05552},
  year={2026}
}

@inproceedings{li2026modeling,
  title={Modeling Uncertainty Trends for Timely Retrieval in Dynamic RAG},
  author={Li, Bo and Tian, Tian and Xu, Zhenghua and Cheng, Hao and Zhang, Shikun and Ye, Wei},
  booktitle={Proceedings of the AAAI Conference on Artificial Intelligence},
  volume={40},
  number={37},
  pages={31527--31535},
  year={2026}
}

@article{zeng2026vision,
  title={Vision-deepresearch benchmark: Rethinking visual and textual search for multimodal large language models},
  author={Yu Zeng and Wenxuan Huang and Zhen Fang and Shuang Chen and Yufan Shen and Yishuo Cai and Xiaoman Wang and Zhenfei Yin and Lin Chen and Zehui Chen and Shiting Huang and Yiming Zhao and Xu Tang and Yao Hu and Philip Torr and Wanli Ouyang and Shaosheng Cao},
  journal={arXiv preprint arXiv:2602.02185},
  year={2026}
}

@inproceedings{huang-etal-2025-critictool,
    title = "{CRITICTOOL}: Evaluating Self-Critique Capabilities of Large Language Models in Tool-Calling Error Scenarios",
    author = "Huang, Shiting  and
      Fang, Zhen  and
      Chen, Zehui  and
      Yuan, Siyu  and
      Ye, Junjie  and
      Zeng, Yu  and
      Chen, Lin  and
      Mao, Qi  and
      Zhao, Feng",
    booktitle = "Proceedings of the 2025 Conference on Empirical Methods in Natural Language Processing",
    month = nov,
    year = "2025",
    url = "https://aclanthology.org/2025.emnlp-main.1355/",
    doi = "10.18653/v1/2025.emnlp-main.1355",
    pages = "26672--26704",

}

@article{yang2026infact,
  title={INFACT: A Diagnostic Benchmark for Induced Faithfulness and Factuality Hallucinations in Video-LLMs},
  author={Yang, Junqi and Min, Yuecong and Zhang, Jie and Shan, Shiguang and Chen, Xilin},
  journal={arXiv preprint arXiv:2603.11481},
  year={2026}
}

@article{zhou2025flattery,
  title={Flattery in motion: Benchmarking and analyzing sycophancy in video-llms},
  author={Zhou, Wenrui and Hendy, Mohamed and Yang, Shu and Yang, Qingsong and Guo, Zikun and Luo, Yuyu and Hu, Lijie and Wang, Di},
  journal={arXiv preprint arXiv:2506.07180},
  year={2025}
}

@article{zhou2025carpo,
  title={CARPO: Leveraging Listwise Learning-to-Rank for Context-Aware Query Plan Optimization},
  author={Zhou, Wenrui and Liu, Qiyu and Peng, Jingshu and Zhang, Aoqian and Chen, Lei},
  journal={arXiv preprint arXiv:2509.03102},
  year={2025}
}

@article{white2023prompt,
  title={A prompt pattern catalog to enhance prompt engineering with chatgpt},
  author={White, Jules and Fu, Quchen and Hays, Sam and Sandborn, Michael and Olea, Carlos and Gilbert, Henry and Elnashar, Ashraf and Spencer-Smith, Jesse and Schmidt, Douglas C},
  journal={arXiv preprint arXiv:2302.11382},
  year={2023}
}

@misc{li2025cama,
      title={Make LVLMs Focus: Context-Aware Attention Modulation for Better Multimodal In-Context Learning}, 
      author={Yanshu Li and Jianjiang Yang and Ziteng Yang and Bozheng Li and Ligong Han and Hongyang He and Zhengtao Yao and Yingjie Victor Chen and Songlin Fei and Dongfang Liu and Ruixiang Tang},
      year={2025},
      eprint={2505.17097},
      archivePrefix={arXiv},
      primaryClass={cs.CV},
      url={https://arxiv.org/abs/2505.17097}, 
}

@misc{li2025towards,
      title={Towards Generalizable Implicit In-Context Learning with Attention Routing}, 
      author={Jiaqian Li and Yanshu Li and Ligong Han and Ruixiang Tang and Wenya Wang},
      year={2025},
      eprint={2509.22854},
      archivePrefix={arXiv},
      primaryClass={cs.CL},
      url={https://arxiv.org/abs/2509.22854}, 
}

@article{xiao2026not,
  title={Not All Directions Matter: Toward Structured and Task-Aware Low-Rank Adaptation},
  author={Xiao, Xi and Ma, Chenrui and Zhang, Yunbei and Liu, Chen and Wang, Zhuxuanzi and Li, Yanshu and Zhao, Lin and Hu, Guosheng and Wang, Tianyang and Xu, Hao},
  journal={arXiv preprint arXiv:2603.14228},
  year={2026}
}

@misc{zhang2026expseek,
      title={ExpSeek: Self-Triggered Experience Seeking for Web Agents}, 
      author={Wenyuan Zhang and Xinghua Zhang and Haiyang Yu and Shuaiyi Nie and Bingli Wu and Juwei Yue and Tingwen Liu and Yongbin Li},
      year={2026},
      eprint={2601.08605},
      archivePrefix={arXiv},
      primaryClass={cs.CL},
      url={https://arxiv.org/abs/2601.08605}, 
}

@inproceedings{zhang-etal-2025-sotopia,
    title = "{SOTOPIA}-{\ensuremath{\Omega}}: Dynamic Strategy Injection Learning and Social Instruction Following Evaluation for Social Agents",
    author = "Zhang, Wenyuan  and
      Liu, Tianyun  and
      Song, Mengxiao  and
      Li, Xiaodong  and
      Liu, Tingwen",
    booktitle = "Proceedings of the 63rd Annual Meeting of the Association for Computational Linguistics (Volume 1: Long Papers)",
    year = "2025",
    url = "https://aclanthology.org/2025.acl-long.1203/",
    doi = "10.18653/v1/2025.acl-long.1203",
    pages = "24669--24697",
    
}

@article{zhang2025s1,
  title={S1-bench: A simple benchmark for evaluating system 1 thinking capability of large reasoning models},
  author={Zhang, Wenyuan and Nie, Shuaiyi and Zhang, Xinghua and Zhang, Zefeng and Liu, Tingwen},
  journal={arXiv preprint arXiv:2504.10368},
  year={2025}
}

@misc{lin2025causal2vec,
      title={Causal2Vec: Improving Decoder-only LLMs as Versatile Embedding Models}, 
      author={Ailiang Lin and Zhuoyun Li and Kotaro Funakoshi and Manabu Okumura},
      year={2025},
      eprint={2507.23386},
      archivePrefix={arXiv},
      primaryClass={cs.CL},
      url={https://arxiv.org/abs/2507.23386}, 
}

@article{egressy2025set,
  title={Set-llm: A permutation-invariant llm},
  author={Egressy, Beni and St{\"u}hmer, Jan},
  journal={arXiv preprint arXiv:2505.15433},
  year={2025}
}

@article{zhang2024trendfact,
  title={TrendFact: A Benchmark for Explainable Hotspot Perception in Fact-Checking with Natural Language Explanation},
  author={Zhang, Xiaocheng and Wang, Xi and Lu, Yifei and Wang, Jianing and Ye, Zhuangzhuang and Bao, Mengjiao and Yan, Peng and Su, Xiaohong},
  journal={arXiv preprint arXiv:2410.15135},
  year={2024}
}

@inproceedings{liu-etal-2021-improving-embedding-based,
    title = "Improving Embedding-based Large-scale Retrieval via Label Enhancement",
    author = "Liu, Peiyang  and
      Wang, Xi  and
      Wang, Sen  and
      Ye, Wei  and
      Xi, Xiangyu  and
      Zhang, Shikun",
    booktitle = "Findings of the Association for Computational Linguistics: EMNLP 2021",
    year = "2021",
    url = "https://aclanthology.org/2021.findings-emnlp.13/",
    doi = "10.18653/v1/2021.findings-emnlp.13",
    pages = "133--142",
}

@inproceedings{liu2025queries,
  title={Queries Are Not Alone: Clustering Text Embeddings for Video Search},
  author={Liu, Peiyang and Wang, Xi and Cui, Ziqiang and Ye, Wei},
  booktitle={Proceedings of the 48th International ACM SIGIR Conference on Research and Development in Information Retrieval},
  pages={874--883},
  year={2025}
}

@inproceedings{liu2021distilling,
  title={Distilling knowledge from bert into simple fully connected neural networks for efficient vertical retrieval},
  author={Liu, Peiyang and Wang, Xi and Wang, Lin and Ye, Wei and Xi, Xiangyu and Zhang, Shikun},
  booktitle={Proceedings of the 30th ACM International Conference on Information \& Knowledge Management},
  pages={3965--3975},
  year={2021}
}

@inproceedings{yuan2026strucsum,
    title = "{S}truc{S}um: Graph-Structured Reasoning for Long Document Extractive Summarization with {LLM}s",
    author = "Yuan, Haohan  and
      Hong, Sukhwa  and
      Zhang, Haopeng",
    booktitle = "Findings of the {A}ssociation for {C}omputational {L}inguistics: {EACL} 2026",
    year = "2026",
    url = "https://aclanthology.org/2026.findings-eacl.192/",
    doi = "10.18653/v1/2026.findings-eacl.192",
    pages = "3708--3721",
}

@inproceedings{yuan2025domainsum,
    title = "{D}omain{S}um: A Hierarchical Benchmark for Fine-Grained Domain Shift in Abstractive Text Summarization",
    author = "Yuan, Haohan  and
      Zhang, Haopeng",
    booktitle = "Findings of the Association for Computational Linguistics: NAACL 2025",
    year = "2025",
    url = "https://aclanthology.org/2025.findings-naacl.118/",
    doi = "10.18653/v1/2025.findings-naacl.118",
    pages = "2219--2231",
}

@article{yuan2025understanding,
  title={Understanding LLM Reasoning for Abstractive Summarization},
  author={Yuan, Haohan and Zhang, Haopeng},
  journal={arXiv preprint arXiv:2512.03503},
  year={2025}
}

@inproceedings{ge2026expert,
  title={Expert-Guided Prompting and Retrieval-Augmented Generation for Emergency Medical Service Question Answering},
  author={Ge, Xueren and Murtaza, Sahil and Cortez, Anthony and Alemzadeh, Homa},
  booktitle={Proceedings of the AAAI Conference on Artificial Intelligence},
  volume={40},
  number={36},
  pages={30798--30806},
  year={2026}
}

@article{zhang2026completion,
  title={From Completion to Editing: Unlocking Context-Aware Code Infilling via Search-and-Replace Instruction Tuning},
  author={Jiajun Zhang and Zeyu Cui and Jiaxi Yang and Lei Zhang and Yuheng Jing and Zeyao Ma and Tianyi Bai and Zilei Wang and Qiang Liu and Liang Wang and Binyuan Hui and Junyang Lin},
  journal={arXiv preprint arXiv:2601.13384},
  year={2026}
}

@article{zhang2026realchart2code,
  title={RealChart2Code: Advancing Chart-to-Code Generation with Real Data and Multi-Task Evaluation},
  author={Jiajun Zhang and Yuying Li and Zhixun Li and Xingyu Guo and Jingzhuo Wu and Leqi Zheng and Yiran Yang and Jianke Zhang and Qingbin Li and Shannan Yan and Zhetong Li and Changguo Jia and Junfei Wu and Zilei Wang and Qiang Liu and Liang Wang},
  journal={arXiv preprint arXiv:2603.25804},
  year={2026}
}

@article{wang2026perm,
  title={PERM: Psychology-grounded Empathetic Reward Modeling for Large Language Models},
  author={Wang, Chengbing and Zheng, Wuqiang and Zhang, Yang and Zhu, Fengbin and Cheng, Junyi and Xie, Yi and Wang, Wenjie and Feng, Fuli},
  journal={arXiv preprint arXiv:2601.10532},
  year={2026}
}

@article{wang2025think,
  title={Think-While-Generating: On-the-Fly Reasoning for Personalized Long-Form Generation},
  author={Wang, Chengbing and Zhang, Yang and Wang, Wenjie and Zhao, Xiaoyan and Feng, Fuli and He, Xiangnan and Chua, Tat-Seng},
  journal={arXiv preprint arXiv:2512.06690},
  year={2025}
}

@inproceedings{li-etal-2025-cmie,
    title = "{CMIE}: Combining {MLLM} Insights with External Evidence for Explainable Out-of-Context Misinformation Detection",
    author = "Li, Fanxiao  and
      Wu, Jiaying  and
      He, Canyuan  and
      Zhou, Wei",
    booktitle = "Findings of the Association for Computational Linguistics: ACL 2025",
    year = "2025",
    url = "https://aclanthology.org/2025.findings-acl.487/",
    doi = "10.18653/v1/2025.findings-acl.487",
    pages = "9342--9354",
}

@article{li2026s,
  title={What's Left Unsaid? Detecting and Correcting Misleading Omissions in Multimodal News Previews},
  author={Li, Fanxiao and Wu, Jiaying and Fu, Tingchao and Li, Dayang and Wan, Herun and Zhou, Wei and Kan, Min-Yen},
  journal={arXiv preprint arXiv:2601.05563},
  year={2026}
}

@article{yang2025benchmarking,
  title={Benchmarking multimodal RAG through a chart-based document question-answering generation framework},
  author={Yang, Yuming and Zhong, Jiang and Jin, Li and Huang, Jingwang and Gao, Jingpeng and Liu, Qing and Bai, Yang and Zhang, Jingyuan and Jiang, Rui and Wei, Kaiwen},
  journal={arXiv preprint arXiv:2502.14864},
  year={2025}
}

@inproceedings{wang2026one,
  title={This One or That One? A Study on Accessibility via Demonstratives with Multimodal Large Language Models},
  author={Wang, Yu and Chersoni, Emmanuele and Huang, Chu-Ren},
  booktitle={Language Resources and Evaluation Conference 2026},
  year={2026},
}

@article{zhang2026towards,
  title={Towards Reliable Multimodal Disaster Severity Assessment through Preference Optimization and Explainable Vision-Language Reasoning},
  author={Zhang, Yuanjun and Shaik, Fuzel Ahamed and Acharjee, Suvojit and Khalid, Fahad and Oussalah, Mourad},
  journal={Reliability Engineering \& System Safety},
  pages={112674},
  year={2026},
  publisher={Elsevier}
}

@article{qian2025hyfedrag,
  title={HyFedRAG: A Federated Retrieval-Augmented Generation Framework for Heterogeneous and Privacy-Sensitive Data},
  author={Qian, Cheng and Zhang, Hainan and Tong, Yongxin and Zheng, Hong-Wei and Zheng, Zhiming},
  journal={arXiv preprint arXiv:2509.06444},
  year={2025}
}

@article{jiang2026magma,
  title={MAGMA: A Multi-Graph based Agentic Memory Architecture for AI Agents},
  author={Jiang, Dongming and Li, Yi and Li, Guanpeng and Li, Bingzhe},
  journal={arXiv preprint arXiv:2601.03236},
  year={2026}
}

@article{jiang2026anatomy,
  title={Anatomy of Agentic Memory: Taxonomy and Empirical Analysis of Evaluation and System Limitations},
  author={Dongming Jiang and Yi Li and Songtao Wei and Jinxin Yang and Ayushi Kishore and Alysa Zhao and Dingyi Kang and Xu Hu and Feng Chen and Qiannan Li and Bingzhe Li},
  journal={arXiv preprint arXiv:2602.19320},
  year={2026}
}

@inproceedings{sun-etal-2025-causalabstain,
    title = "{C}ausal{A}bstain: Enhancing Multilingual {LLM}s with Causal Reasoning for Trustworthy Abstention",
    author = "Sun, Yuxi  and
      Zuo, Aoqi  and
      Gao, Wei  and
      Ma, Jing",
    booktitle = "Findings of the Association for Computational Linguistics: ACL 2025",
    year = "2025",
    url = "https://aclanthology.org/2025.findings-acl.723/",
    doi = "10.18653/v1/2025.findings-acl.723",
    pages = "14060--14076",
}

@article{wen-etal-2025-know,
    title = "Know Your Limits: A Survey of Abstention in Large Language Models",
    author = "Wen, Bingbing  and
      Yao, Jihan  and
      Feng, Shangbin  and
      Xu, Chenjun  and
      Tsvetkov, Yulia  and
      Howe, Bill  and
      Wang, Lucy Lu",
    journal = "Transactions of the Association for Computational Linguistics",
    volume = "13",
    year = "2025",
    url = "https://aclanthology.org/2025.tacl-1.26/",
    doi = "10.1162/tacl_a_00754",
    pages = "529--556",
}

@article{fu2026s,
  title={S-Path-RAG: Semantic-Aware Shortest-Path Retrieval Augmented Generation for Multi-Hop Knowledge Graph Question Answering},
  author={Fu, Rong and Wang, Yemin and Xu, Tianxiang and Liu, Yongtai and Tang, Weizhi and Wu, Wangyu and Ma, Xiaowen and Fong, Simon},
  journal={arXiv preprint arXiv:2603.23512},
  year={2026}
}

@article{fu2026neurosymactive,
  title={NeuroSymActive: Differentiable Neural-Symbolic Reasoning with Active Exploration for Knowledge Graph Question Answering},
  author={Rong Fu and Yang Li and Zeyu Zhang and Jiekai Wu and Yaohua Liu and Shuaishuai Cao and Yangchen Zeng and Yuhang Zhang and Xiaojing Du and Chuang Zhao and Kangning Cui and Simon Fong},
  journal={arXiv preprint arXiv:2602.15353},
  year={2026}
}

@article{zhang2025stage,
  title={STAGE: Storyboard-Anchored Generation for Cinematic Multi-shot Narrative},
  author={Zhang, Peixuan and Jia, Zijian and Liu, Kaiqi and Weng, Shuchen and Li, Si and Shi, Boxin},
  journal={arXiv preprint arXiv:2512.12372},
  year={2025}
}

@misc{li2026retrievalgenerationunifiedframework,
      title={Retrieval as Generation: A Unified Framework with Self-Triggered Information Planning}, 
      author={Bo Li and Mingda Wang and Gexiang Fang and Shikun Zhang and Wei Ye},
      year={2026},
      eprint={2604.11407},
      archivePrefix={arXiv},
      primaryClass={cs.CL},
      url={https://arxiv.org/abs/2604.11407}, 
}

@misc{xu2026selfcorrectingragenhancingfaithfulness,
      title={Self-Correcting RAG: Enhancing Faithfulness via MMKP Context Selection and NLI-Guided MCTS}, 
      author={Shijia Xu and Zhou Wu and Xiaolong Jia and Yu Wang and Kai Liu and April Xiaowen Dong},
      year={2026},
      eprint={2604.10734},
      archivePrefix={arXiv},
      primaryClass={cs.CL},
      url={https://arxiv.org/abs/2604.10734}, 
}

@misc{sun2026factecausalityinspiredevaluationtrustworthy,
      title={FACT-E: Causality-Inspired Evaluation for Trustworthy Chain-of-Thought Reasoning}, 
      author={Yuxi Sun and Aoqi Zuo and Haotian Xie and Wei Gao and Mingming Gong and Jing Ma},
      year={2026},
      eprint={2604.10693},
      archivePrefix={arXiv},
      primaryClass={cs.AI},
      url={https://arxiv.org/abs/2604.10693}, 
}

@article{jiang2026cmedtebcarebenchmarking,
  title={CMedTEB \& CARE: Benchmarking and Enabling Efficient Chinese Medical Retrieval via Asymmetric Encoders},
  author={Jiang, Angqing and Chen, Jianlyu and Fang, Zhe and Wang, Yongcan and Li, Xinpeng and Ding, Keyu and Lian, Defu},
  journal={arXiv preprint arXiv:2604.10937},
  year={2026}
}

@misc{kong2026causalgazeunveilinghallucinationscounterfactual,
      title={CausalGaze: Unveiling Hallucinations via Counterfactual Graph Intervention in Large Language Models}, 
      author={Linggang Kong and Lei Wu and Yunlong Zhang and Xiaofeng Zhong and Zhen Wang and Yongjie Wang and Yao Pan},
      year={2026},
      eprint={2604.11087},
      archivePrefix={arXiv},
      primaryClass={cs.LG},
      url={https://arxiv.org/abs/2604.11087}, 
}

@inproceedings{zhu-etal-2024-information,
    title = "An Information Bottleneck Perspective for Effective Noise Filtering on Retrieval-Augmented Generation",
    author = "Zhu, Kun  and
      Feng, Xiaocheng  and
      Du, Xiyuan  and
      Gu, Yuxuan  and
      Yu, Weijiang  and
      Wang, Haotian  and
      Chen, Qianglong  and
      Chu, Zheng  and
      Chen, Jingchang  and
      Qin, Bing",
    booktitle = "Proceedings of the 62nd Annual Meeting of the Association for Computational Linguistics (Volume 1: Long Papers)",
    year = "2024",
    url = "https://aclanthology.org/2024.acl-long.59/",
    doi = "10.18653/v1/2024.acl-long.59",
    pages = "1044--1069",
}

@inproceedings{
hamman2025improving,
title={Improving Consistency in Retrieval-Augmented Systems with Group Similarity Reward},
author={Faisal Hamman and Chenyang Zhu and Anoop Kumar and Xujun Peng and Sanghamitra Dutta and Daben Liu and Alfy Samuel},
booktitle={NeurIPS 2025 Workshop: Reliable ML from Unreliable Data},
year={2025},
url={https://openreview.net/forum?id=He0YGNuM9F}
}

\appendix
\clearpage

\begin{center}
\huge\textbf{Appendix}
\end{center}

\begin{flushleft}
\Large\textbf{Contents}
\rule{\linewidth}{1pt}
\end{flushleft}

\startcontents[sections]
\printcontents[sections]{}{1}{}

\rule{\linewidth}{1pt}

\section{Implementation Details}
\subsection{Datasets}
\label{appendix:datasets}
We conduct experiments on three widely used QA datasets that cover both single-hop and multi-hop question-answering scenarios. Table~\ref{tab:qa-datasets} summarizes the key statistics of these datasets. Specifically, \textbf{NQ}~\cite{nq} and \textbf{TriviaQA}~\cite{triviaqa} are representative single-hop datasets, where each question can typically be answered using information from a single passage retrieved from the corpus. These datasets primarily evaluate a model’s ability to locate and extract factual evidence efficiently. In contrast, \textbf{HotpotQA}~\cite{hotpotqa} is a challenging multi-hop dataset that requires integrating and reasoning over multiple pieces of evidence distributed across different documents to derive the final answer. This dataset is particularly useful for testing a model’s reasoning and compositional understanding capabilities. 
Together, these datasets provide a comprehensive benchmark for evaluating both the retrieval quality and reasoning robustness of our proposed method under diverse task settings.

\begin{table}[h]
\small
\centering
\begin{tabular}{lcccc}
\toprule[1.1pt]
\textbf{Dataset} & \textbf{Type}&\textbf{\# Train} & \textbf{\# Dev} & \textbf{\# Test} \\
\midrule
NQ      &  single-hop & 79.1k  & 8.7k  & 3.6k  \\
TriviaQA &  single-hop& 78.7k  & 11.3k  & 8.8k \\
HotpotQA  & multi-hop& 88.9k  & 5.6k  & 5.6k  \\
\bottomrule[1.1pt]
\end{tabular}
\caption{Statistics for the datasets.}
\label{tab:qa-datasets}
\end{table}

\subsection{Baseline Details}
\label{appendix:baseline}
We compare Stable-RAG with the following baseline strategies. To ensure a fair comparison, all methods are evaluated on the same test set and retrieved set.

\paragraph{Vanilla Methods.}

(i) \textit{Direct Generation}. This baseline relies solely on the generator’s parametric knowledge to produce answers without consulting any retrieved documents.
(ii) \textit{Vanilla RAG}~\cite{lewis2020retrieval}. This baseline concatenates all retrieved documents as model input without any additional processing.
(iii) \textit{Vanilla SFT}. We implement vanilla SFT following ~\citet{zhang-etal-2024-r}. For each training example, this baseline uses the gold answer as the training label if it appears in the retrieved documents; otherwise, it assigns \emph{``I don’t know''} as the training label to guide the model to abstain when the necessary information is missing.

\paragraph{Robust RAG.}
(i) \textit{RetRobust}~\cite{retrobust}. This baseline improves retrieval-augmented QA models by filtering out irrelevant retrieved passages and fine-tuning the model on a mix of relevant and irrelevant contexts, enabling it to leverage relevant information while remaining robust to irrelevant content.
(ii) \textit{ATM}~\cite{zhu-etal-2024-atm}. This baseline optimizes a retrieval-augmented Generator using an Adversarial Tuning Multi-agent system, where an auxiliary Attacker agent iteratively steers the Generator to better discriminate useful documents from noisy or fabricated ones, improving robustness and performance on knowledge-intensive question answering tasks.
(iii) \textit{RAAT}~\cite{raat}. This baseline dynamically adjusts the model’s learning process in response to various types of retrieval noise through adaptive adversarial training, while employing multi-task learning to enable the model to internally recognize and handle noisy contexts, thereby improving robustness and answer quality in retrieval-augmented generation.

\paragraph{Positional Bias.}
(i) \textit{Pos2Distill}~\cite{wang-etal-2025-position}. This baseline mitigates positional bias in long-context tasks by transferring knowledge from advantageous positions to less favorable ones through position-to-position knowledge distillation.
(ii) \textit{Ms-PoE}~\cite{mspoe}. This baseline uses Multi-scale Positional Encoding to mitigate the "lost-in-the-middle" issue in LLMs by rescaling positional indices and assigning different scaling ratios to attention heads, enabling multi-scale context fusion without fine-tuning or extra overhead.

\subsection{Training Details}
\label{appendix:training}

We use LLaMA3-8B-Instruct~\footnote{\url{https://huggingface.co/meta-llama/Meta-Llama-3-8B-Instruct}}~\cite{llama3} and Qwen3-8B~\footnote{\url{https://huggingface.co/Qwen/Qwen3-8B}}~\cite{qwen3} as backbone models following prior and concurrent work~\cite{zhang2025s1,zhang-etal-2025-sotopia,lin2025causal2vec,huang-etal-2025-critictool,yuan2025domainsum,zhang2026expseek,li2025cama,li2025towards}. These models and their variants have been widely used across a variety of tasks and applications~\cite{zhang2025stage,yang2025benchmarking,li-etal-2025-cmie,li2026s,zhang2026completion,wang2026one,zhang2026realchart2code,wang2025think,wang2026perm,jiang2026anatomy,jiang2026magma,fu2026s,fu2026neurosymactive,sun2026factecausalityinspiredevaluationtrustworthy,jiang2026cmedtebcarebenchmarking,zeng2026vision,ma2026cast}.
We implement DPO training pipeline using the HuggingFace Transformers~\cite{wolf-etal-2020-transformers}, incorporating PEFT LoRA~\cite{hu2022lora,xiao2026not} for parameter-efficient fine-tuning. 
Both the base model and reference model are initialized from pre-trained checkpoints, with the reference model kept in evaluation mode to provide stable policy targets during training.
Each dataset is randomly shuffled and split into 85\% training and 15\% validation samples, with a maximum of 18,000 samples per dataset to control computational overhead. We fix the random seed to 42 to ensure reproducibility.
LoRA is applied to all projection layers with rank $r=128$, alpha $=128$, dropout $=0$ and no additional bias terms.
The DPO configurations~\cite{zhang2026towards} use a per-device batch size of 2 with gradient accumulation of 8, a learning rate of $5\times10^{-6}$, a linear warmup ratio of 0.1, and a preference scaling hyperparameter $\beta$ of 0.4. We train LLaMA-3-8B-Instruct for 1 epoch and Qwen3-8B for 2 epochs on two NVIDIA RTX PRO 6000 GPUs, with each epoch taking roughly two hours. Notably, we use greedy decoding during data construction, and set the temperature to 0.01 during inference, which is nearly equivalent to greedy decoding. This ensures that output variations primarily reflect document-order sensitivity rather than sampling randomness.

\subsection{Prompts}
We adopt a system-user style prompting scheme~\cite{white2023prompt,zhou2025flattery,zhou2025carpo} to guide the backbone LLMs to generate concise, document-grounded answers, as presented in Table~\ref{tab:prompt}.

\begin{table}[t]
\begin{tcolorbox}
\verb|<system>|\\
You are a helpful, respectful, and honest assistant. Answer the question with couple of words using the provided documents. For example: Question: What is the capital of France? Output: Paris.\\
\verb|</system>|
\\
\verb|<user>|\\
Question: \verb|{query}|\\
Documents: \\
Doc1: \verb|{Document 1}|  \\
Doc2: \verb|{Document 2}|  \\
...... \\
\verb|</user>|\\
\end{tcolorbox}
\caption{Prompt for the backbone LLMs.}
\label{tab:prompt}
\end{table}

\section{Mathematical Derivations}
\label{math}
We employ spectral clustering on hidden states to identify dominant reasoning modes across permutations of retrieved documents. Compared with conventional clustering methods, spectral clustering captures the global structure of the hidden state space. This enables Stable-RAG to robustly group similar reasoning behaviors, reduce noise from spurious variations, and improve the consistency of preference signals used for DPO alignment.

\subsection{Spectral Clustering on Hidden States}

Spectral clustering is applied to the hidden states matrix:
\[
H = [h^{(1)}, h^{(2)}, \dots, h^{(N)}]^\top \in \mathbb{R}^{N \times d}
\] 
to adaptively determine the number of clusters and capture the global structure of the hidden state space, where each cluster corresponds to a latent reasoning mode~\cite{lee-etal-2025-efficient-latent}.

\subsection{Similarity Graph and Adjacency Matrix}

We construct a weighted similarity graph \(G = (V, E)\) where each node corresponds to a hidden state \(h^{(i)}\) and edges encode pairwise similarities. The adjacency matrix \(A \in \mathbb{R}^{N \times N}\) is computed as the exponential of the cosine distance:
\[
A_{ij} = \exp\Big(-\frac{1 - \frac{h^{(i)} \cdot h^{(j)}}{\|h^{(i)}\| \, \|h^{(j)}\|}}{\sigma}\Big),
\]
where \(\sigma\) is a hyperparameter controlling sensitivity.

\subsection{Degree Matrix and Normalized Laplacian}

The degree matrix \(D\) is a diagonal matrix with entries:
\[
D_{ii} = \sum_{j=1}^{N} A_{ij}.
\]

The normalized graph Laplacian is:
\[
L = I - D^{-1/2} A D^{-1/2},
\]
where \(I\) is the identity matrix.

\subsection{Eigen-decomposition and Determining Cluster Number}

Let \(\lambda_1 \le \dots \le \lambda_N\) be the eigenvalues of \(L\). Define the consecutive eigengaps as:
\[
\mathrm{gap}_i = \lambda_{i+1} - \lambda_i.
\]

The number of clusters \(K\) is set adaptively as:
\[
K = \max\bigl(2, (\arg\max_i \mathrm{gap}_i) + 1\bigr),
\]
ensuring clear separation between latent reasoning modes following standard practice~\cite{ng2001spectral,von2007tutorial}.

\subsection{Spectral Embedding and Clustering}

We then compute the first \(K\) eigenvectors of \(L\), normalize each row to unit length, and apply standard clustering to assign each hidden state \(h^{(i)}\) to one of the clusters:
\[
C_1, C_2, \dots, C_K,
\] 
exactly following the procedure described in the main text.

\section{More Experimental Results}

\begin{figure}[t!]
\centering
  \includegraphics[width=1.0\columnwidth]{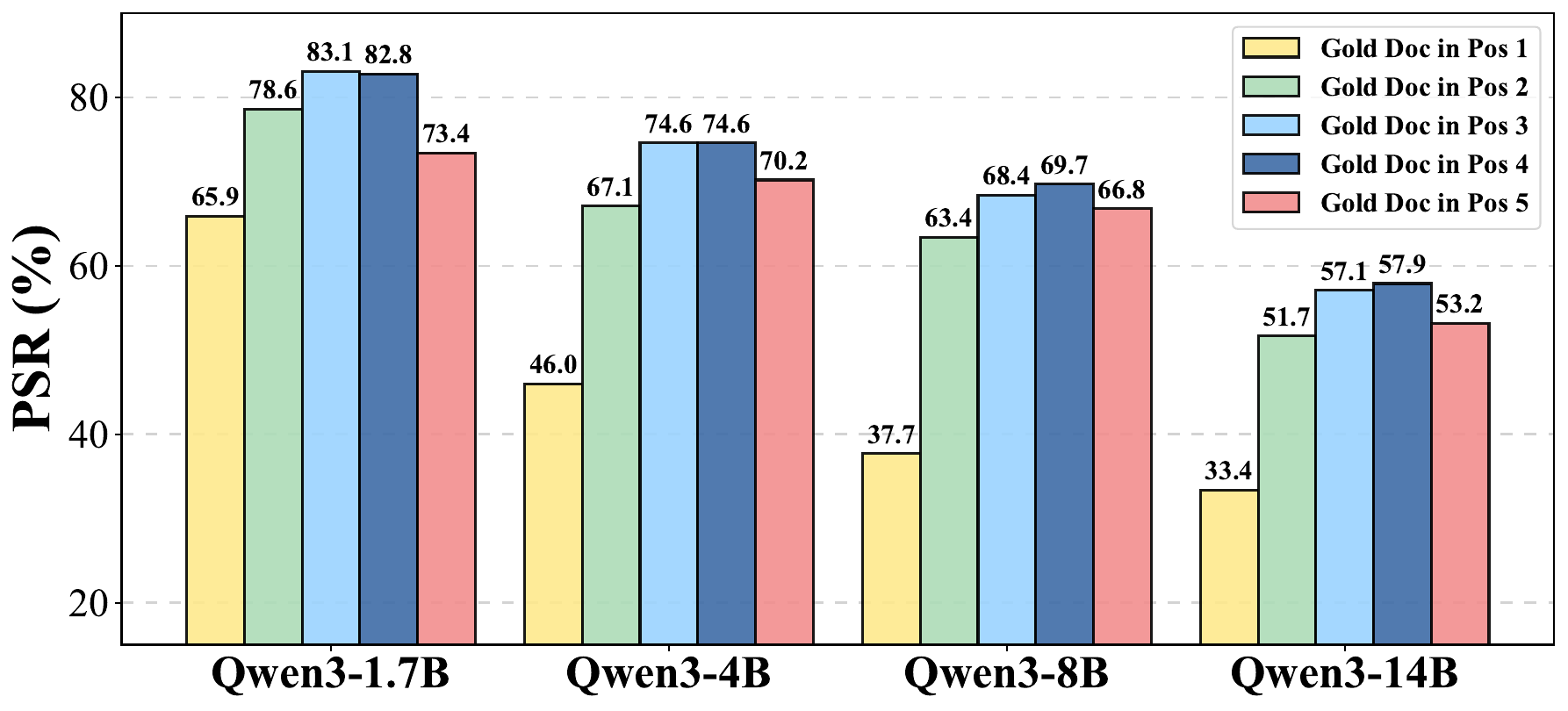}
\caption{\textbf{Perturbation Success Rate (PSR)} on the NQ test set across different Qwen3 models. PSR is computed as the proportion of successful document-order perturbations to produce hallucination results among 1,000 randomly sampled instances, with the gold document fixed in different positions.}
  \label{fig:motivation_qwen3}
\end{figure}

\begin{figure*}[!t]
    \centering

        \begin{subfigure}{0.24\linewidth}
            \includegraphics[width=\linewidth]{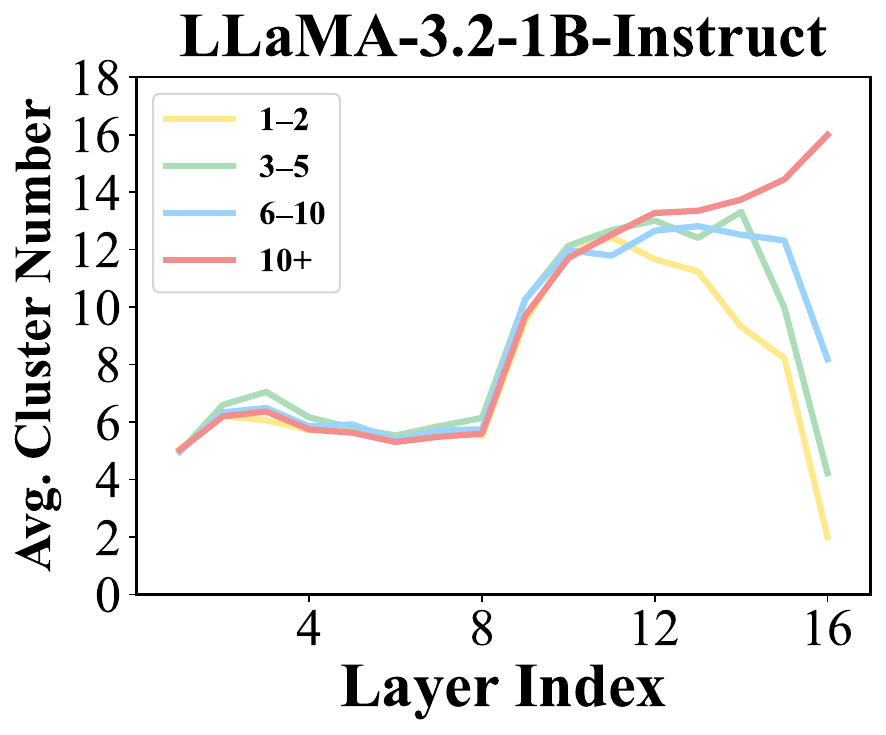}
        \end{subfigure}
        \hfill
        \begin{subfigure}{0.24\linewidth}
            \includegraphics[width=\linewidth]{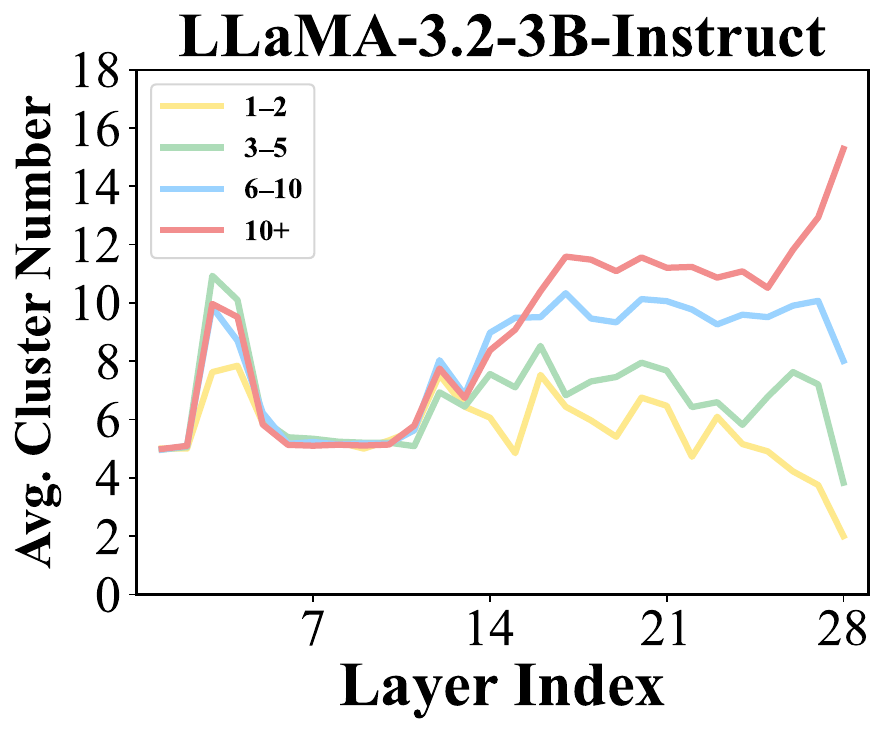}
        \end{subfigure}
        \hfill
        \begin{subfigure}{0.24\linewidth}
            \includegraphics[width=\linewidth]{figures/motivation2_llama3-8b.pdf}
        \end{subfigure}
        \begin{subfigure}{0.24\linewidth}
            \includegraphics[width=\linewidth]{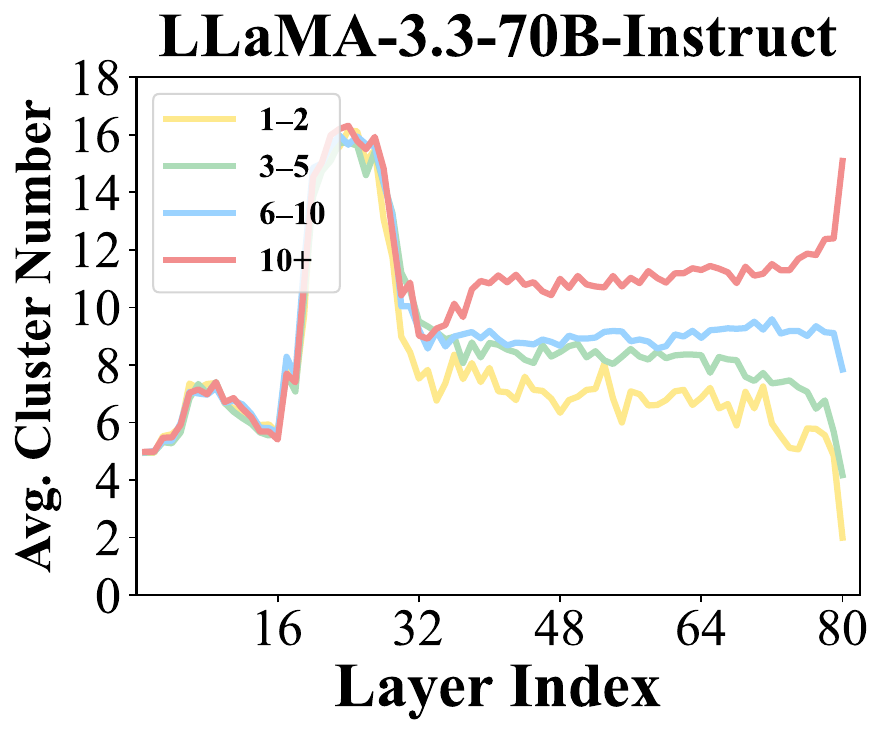}
        \end{subfigure}

        \begin{subfigure}{0.24\linewidth}
        \includegraphics[width=\linewidth]{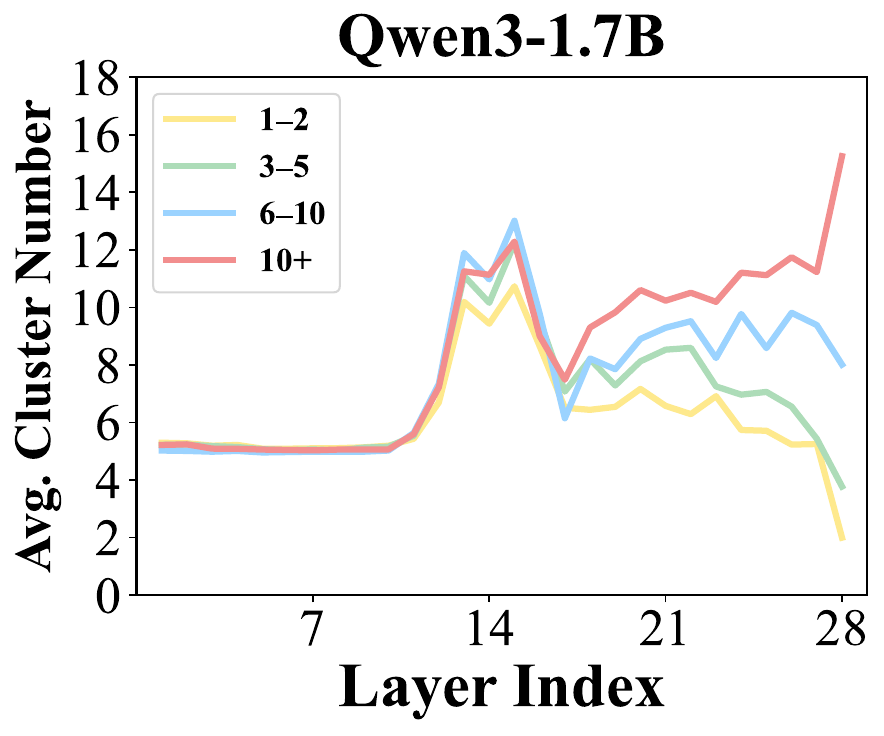}
        % \subcaption{}
        \label{}
    \end{subfigure}
    \hfill
    \begin{subfigure}{0.24\linewidth}
        \includegraphics[width=\linewidth]{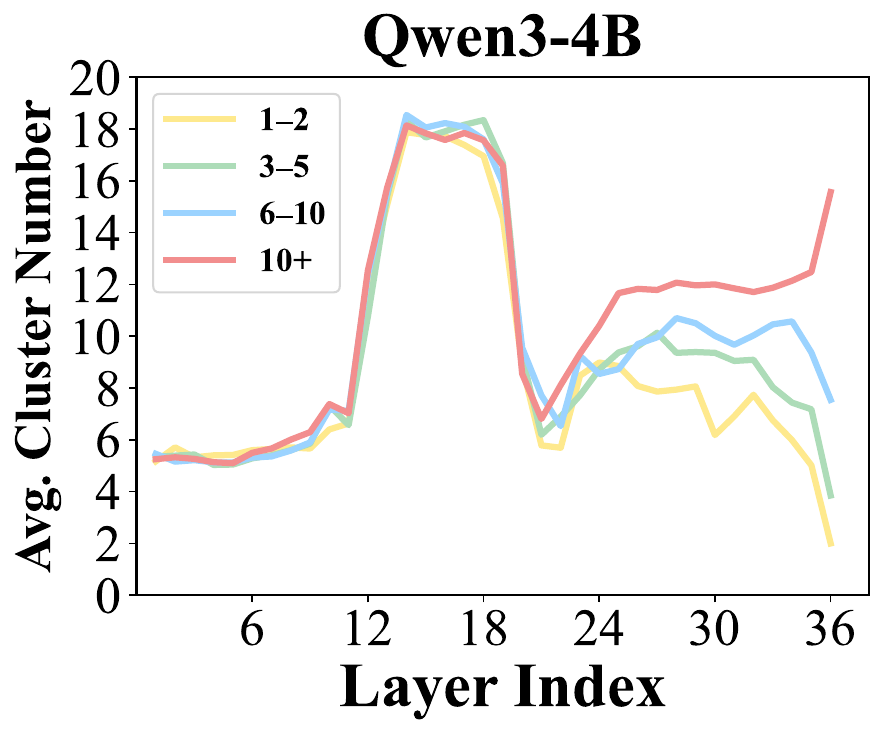}
        % \subcaption{}
        \label{}
    \end{subfigure}
    % \hfill
 \begin{subfigure}{0.24\linewidth}
        \includegraphics[width=\linewidth]{figures/motivation2_qwen3-8b.pdf}
        % \subcaption{}
        \label{}
    \end{subfigure}
     \hfill
 \begin{subfigure}{0.24\linewidth}
        \includegraphics[width=\linewidth]{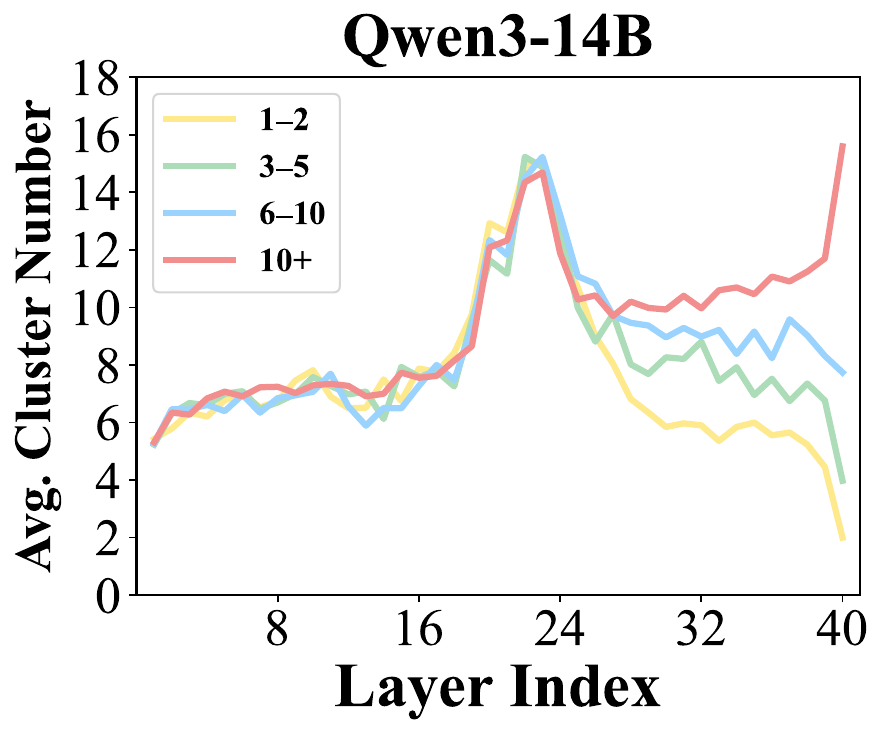}
        % \subcaption{}
        \label{}
    \end{subfigure}

    \caption{Hidden-state clustering behaviors across layers for LLaMA3 series on the NQ train set with DPR retriever and Qwen3 series on the HotpotQA train set with Contriever retriever, using 1,000 randomly sampled instances. Different colored lines indicate the number of clusters of final reasoning states produced by the LLM under all $5! (=120)$ permutations of the Top-5 retrieved documents (e.g., the green line indicates 3–5 cluster states).}
    \label{fig:llama3_cluster}
  
\end{figure*}

% \begin{figure*}[!t]
%     \centering
%     \begin{subfigure}{0.24\linewidth}
%         \includegraphics[width=\linewidth]{motivation2_qwen3-1_7b.pdf}
%         % \subcaption{}
%         \label{}
%     \end{subfigure}
%     \hfill
%     \begin{subfigure}{0.24\linewidth}
%         \includegraphics[width=\linewidth]{motivation2_qwen3-4b.pdf}
%         % \subcaption{}
%         \label{}
%     \end{subfigure}
%     % \hfill
%  \begin{subfigure}{0.24\linewidth}
%         \includegraphics[width=\linewidth]{motivation2_qwen3-8b.pdf}
%         % \subcaption{}
%         \label{}
%     \end{subfigure}
%      \hfill
%  \begin{subfigure}{0.24\linewidth}
%         \includegraphics[width=\linewidth]{motivation2_qwen3-14b.pdf}
%         % \subcaption{}
%         \label{}
%     \end{subfigure}
    
%   \caption{Hidden-state clustering behaviors across layers for Qwen3 series models on the HotpotQA train set with Contriever retriever, using 1,000 random sampled instances. Different colored lines indicate the number of clusters of final reasoning states produced by the LLM under all $5! (=120)$ permutations of the Top-5 retrieved documents (e.g., the green line indicates 3–5 cluster states).}
%      \label{fig:qwen3_cluster}
% \end{figure*}

\subsection{Permutation Sensitivity in Qwen3 Models}
\label{Qwen3}

We further investigate whether document-order sensitivity generalizes to different model families by reporting PSR results on the Qwen3 series. Following the same evaluation protocol as in Figure~\ref{fig:motivation}, we fix the gold document in different positions and measure the proportion of document-order perturbations that lead to hallucinated outputs over 1,000 randomly sampled instances on the NQ test set.

Figure~\ref{fig:motivation_qwen3} compares the PSR trends of the Qwen3 models with those observed in the LLaMA3 Instruct series. Overall, Qwen3 models exhibit clear document-order sensitivity across all model sizes. When the gold document is placed at early positions, the PSR is relatively low, indicating stronger robustness to document-order perturbations. However, as the gold document is shifted to later positions, PSR increases substantially, suggesting a higher likelihood of hallucinations induced purely by document reordering.

We observe a consistent monotonic pattern across Qwen3 variants: PSR generally rises from Top-1 to Top-3 or Top-4 and slightly saturates or declines afterward. This behavior closely mirrors the trends observed in LLaMA3 models, despite differences in model architecture and pretraining data. Moreover, smaller Qwen3 models tend to exhibit higher sensitivity to document order changes, while larger models demonstrate comparatively improved robustness, though the issue remains non-negligible even at larger scales.

These results show that document-order sensitivity is a general property of modern LLMs, highlighting the need for order-robust RAG methods.

\subsection{Structural Instability Across Model Families}
\label{Instability}
We provide additional visualizations of the structural instability in internal reasoning dynamics for both the LLaMA3 and Qwen3 model families as shown in Figure~\ref{fig:llama3_cluster}. We analyze how document permutations induce representation divergence across layers. Despite differences in architecture, scale, and pretraining data, both model families show consistent structural instability. Specifically, shallow-layer representations remain relatively concentrated under document permutations, while strong divergence emerges in middle layers and becomes more pronounced in higher layers. Moreover, high-sensitivity samples consistently exhibit greater representational divergence than stable ones.

These observations suggest that permutation sensitivity originates from a shared structural instability in the reasoning dynamics of large language models rather than from model-specific design choices~\cite{wang2026fbs}. Consistent trends across LLaMA3 and Qwen3 further highlight the need to address structural instability to improve RAG robustness.

\subsection{ Visualization of Layer-wise Hidden States}
\label{app:all_layer}
\begin{figure*}[!h]
  \centering
  \includegraphics[width=\textwidth]{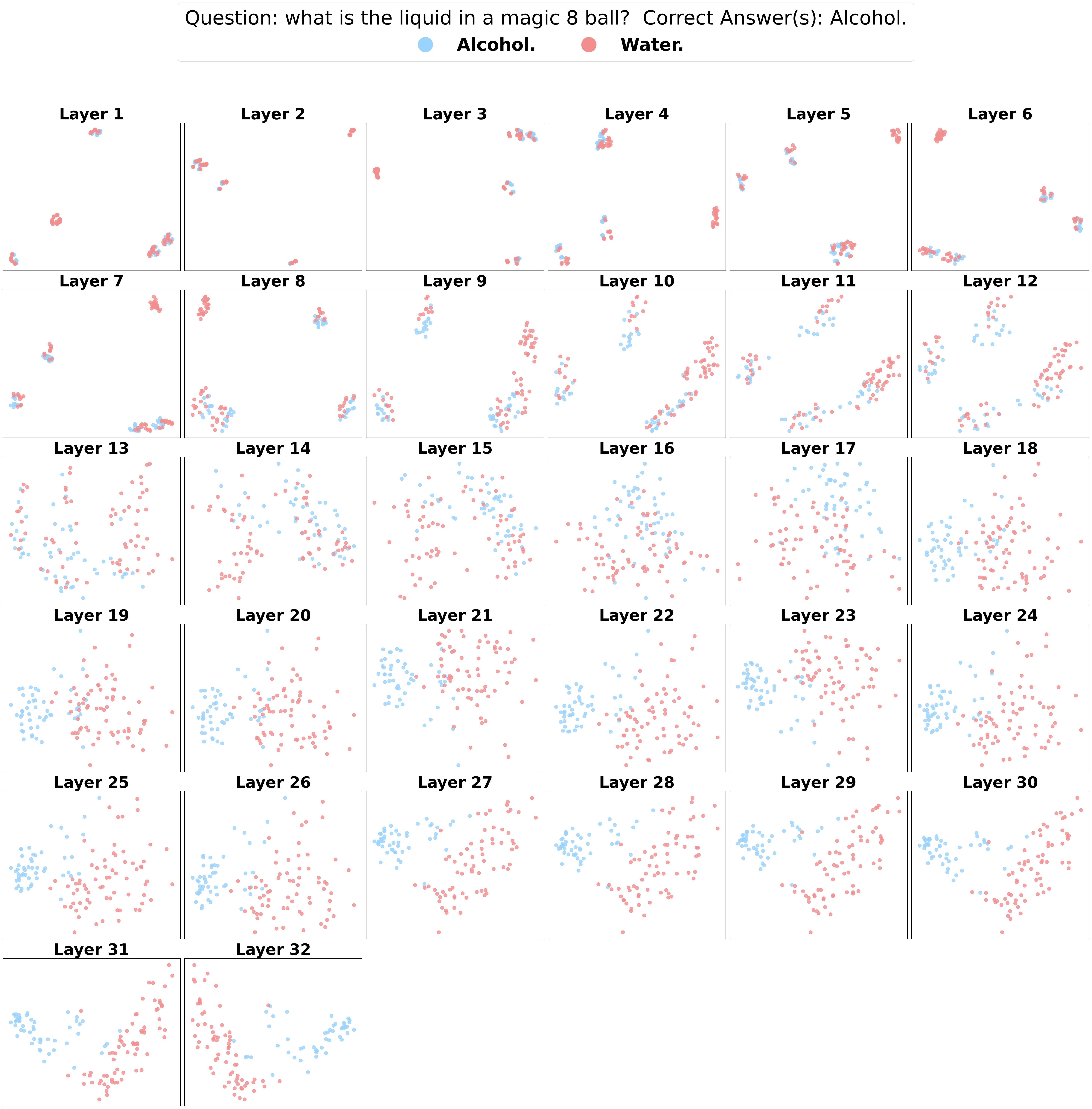}
\caption{2D PCA visualization of hidden state representations across all layers in LLaMA3-8B-Instruct for a single example. Each point corresponds to a document order, and its color represents the model’s final answer.}
  \label{fig:llama3_all_layer}
\end{figure*}

\begin{figure*}[!t]
  \centering
  \includegraphics[width=\textwidth]{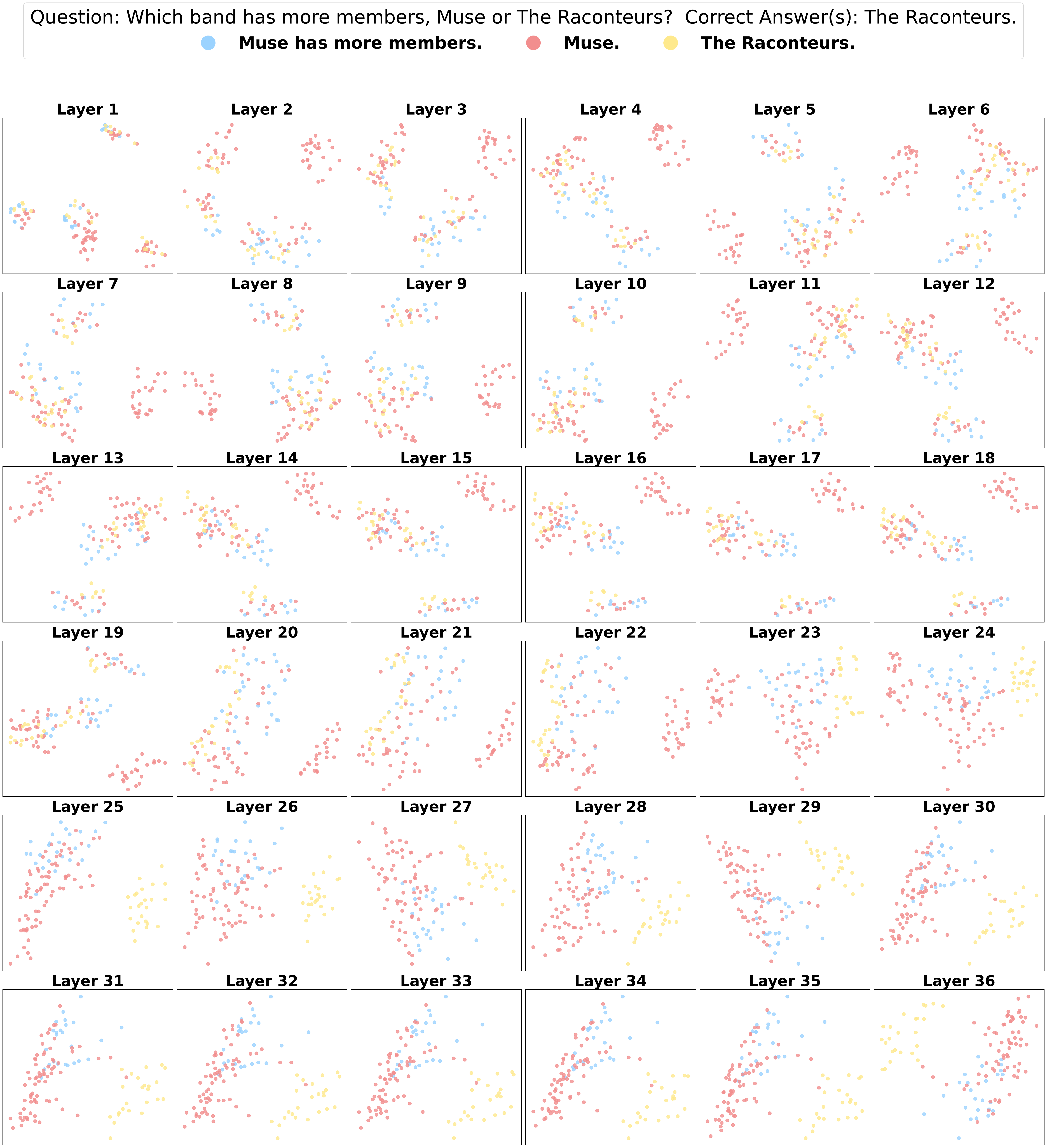}
\caption{2D PCA visualization of hidden state representations across all layers in Qwen3-8B for a single example. Each point corresponds to a document order,
and its color represents the model’s final answer.}
  \label{fig:qwen3_all_layer}
\end{figure*}

Figure~\ref{fig:llama3_all_layer} shows LLaMA3-8B-Instruct on NQ using the Contriever retriever, illustrating the hidden state evolution across all layers for a selected example. Figure~\ref{fig:qwen3_all_layer} displays Qwen3-8B on HotpotQA dataset using Contriever dataset, showing the layer-wise progression of hidden states for a representative sample. 
In both cases, shallow layers exhibit mixed clusters with points corresponding to different answers interleaved, while deeper layers form increasingly well-separated clusters according to the final answers. These visualizations reinforce that the structural evolution of reasoning trajectories is consistent across multiple models and datasets.

\begin{figure}[!t]
\centering
  \includegraphics[width=0.8\columnwidth]{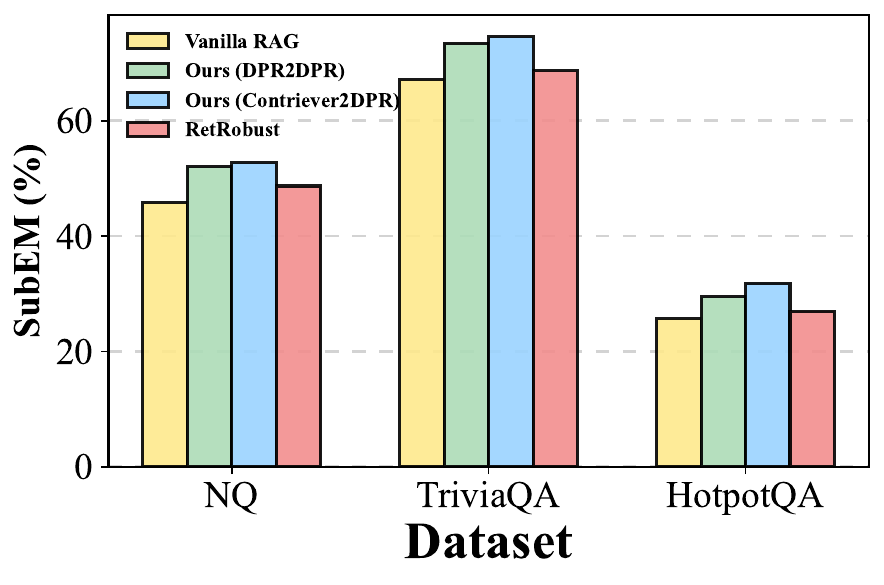}
\caption{Cross-Retriever Transferability.}
  \label{fig:contriever2dpr}
\end{figure}

\subsection{Cross-Retriever Transferability}
\label{app:across-retriever}
As shown in Section~\ref{analysis}, we evaluate transferability from DPR to Contriever. We additionally test transferability from Contriever to DPR in Figure~\ref{fig:contriever2dpr}. Both experiments confirm that Stable-RAG consistently improves answer consistency and reduces permutation-induced variance across retrievers, demonstrating cross-retriever transferability.

\end{document}